\documentclass[sigconf]{acmart}
\makeatletter                   %1
\def\mdseries@tt{m}             %1
\makeatother                    %1
\usepackage[plain]{fancyref}
\usepackage[draft=true]{minted} %2
\usepackage{color}
\usepackage{hyperref}           %5
\hypersetup{
    colorlinks=true,
    linkcolor=blue,
    filecolor=red,      
    urlcolor=magenta,
    breaklinks=true,            %3
}
\usepackage{breakurl}           %3

\begin{document}
\sloppy                         %4

\title{Meta-learning optimizes predictions of missing links in real-world networks}

\author{Bisman Singh}
\affiliation{%
  \institution{Department of Applied Mathematics, University of Colorado}
  \city{Boulder}
  \state{Colorado}
  \country{USA}
}
%\email{bisman.singh@colorado.edu}

\author{Lucy Van Kleunen}
\affiliation{%
  \institution{BioFrontiers Institute, \\ University of Colorado}
  \city{Boulder}
  \state{Colorado}
  \country{USA}}
%\email{lucy.vankleunen@colorado.edu}

\author{Aaron Clauset}
\affiliation{%
 \institution{Department of Computer Science \& BioFrontiers Institute, \\ University of Colorado}
 \city{Boulder}
 \state{Colorado}
 \country{USA}}
%\email{aaron.clauset@colorado.edu}

\renewcommand{\shortauthors}{Singh et al.}

\begin{abstract}
Relational data are ubiquitous in real-world data applications, e.g., in social network analysis or biological modeling, but networks are nearly always incompletely observed. The state-of-the-art for predicting missing links in the hard case of a network without node attributes uses model stacking or neural network techniques. It remains unknown which approach is best, and whether or how the best choice of algorithm depends on the input network's characteristics. We answer these questions systematically using a large, structurally diverse benchmark of 550 real-world networks under two standard accuracy measures (AUC and Top-k), comparing four stacking algorithms with 42 topological link predictors, two of which we introduce here, and two graph neural network algorithms. We show that no algorithm is best across all input networks, all algorithms perform well on most social networks, and few perform well on economic and biological networks. Overall, model stacking with a random forest is both highly scalable and surpasses on AUC or is competitive with graph neural networks on Top-k accuracy. But, algorithm performance depends strongly on network characteristics like the degree distribution, triangle density, and degree assortativity. We introduce a meta-learning algorithm that exploits this variability to optimize link predictions for individual networks by selecting the best algorithm to apply, which we show outperforms all state-of-the-art algorithms and scales to large networks.
\end{abstract}


\received{\today}

\maketitle

\section{Introduction}
Whether links represent interactions among people, as in online social networks, or regulation among genes, as in biological models, relational data or networks are a ubiquitous form of data from complex systems~\cite{newman:networks:2018} and they play an integral role in understanding system structure and predicting system dynamics. However, nearly all real-world networks are incomplete, and predicting the links that are missing is fundamental problem. Missing links can arise from restrictions in access to data~\cite{guimera2009missing}, privacy concerns~\cite{zheleva2009privacy}, e.g., in online settings, the cost of measurement~\cite{zheleva2009privacy}, e.g., in biological systems, or because a network evolves over time~\cite{liben2007link}.

The Link Prediction (LP) problem is to infer these missing (or future) links from patterns of connections in a partially observed network~\cite{cai2014link}. Link prediction has many uses. In evolving networks, it makes educated guesses of future interactions, e.g., new collaborations in an academic coauthorship network~\cite{Huang,wang2014link}, akin to recommendation systems. As a form of cross-validation for networks~\cite{Guimerà}, it can be used to compare different network models, in which better models make more accurate predictions of missing links~\cite{ghasemian2019evaluating}. And, when individual measurements are costly, e.g., in biological systems like protein interactions or ecological systems like food webs, where experimentation is needed to measure whether an interaction exists or not~\cite{han2005modularity}, or even in social systems like public health surveillance or national security, where exhaustive monitoring can be impractical or unethical, link prediction can help manage scarce resources for mapping out a network's structure~\cite{al2011link,vanunu2010predicting}.

The hardest and most general form of the Link Prediction problem makes predictions using only observed links, without leveraging any node attributes or metadata~\cite{ghasemian2019evaluating,ghasemian2020stacking} in the form of scalar and categorical values attached to each node that correlate with the existence of links, e.g., user demographics, species traits, or biological functions. When available, node attributes can substantially enhance the performance of link prediction methods~\cite{vankleunen2024traits,martinez2017survey}, even in ultra-sparse networks with very few edges available for analysis. The utility of node attributes in the Link Prediction problem depends on the causal relationship between a link existing and the attributes of the node pair it connects~\cite{lu2011link}, i.e., $\Pr((i,j)\in E\,|\,f(x_i,x_j))$, where $f()$ is a function of $i$ and $j$'s node attributes~$x_i$ and~$x_j$. If there is no correlation between attributes $x_i$,~$x_j$ and the existence of the link~$(i,j)$, then using node attributes will not improve prediction accuracy for missing links~\cite{abbasi2015critical}. Hence, the hardest form of the Link Prediction problem is precisely when node attributes are not correlated with links or when node attributes are absent entirely. This hard case is the problem setting we consider here, evaluating state-of-the-art algorithms on a broad network benchmark under multiple measures of accuracy.

%% SECTION: RELATED WORK
\section{Related Work}
The literature on the Link Prediction (LP) problem spans multiple fields, including machine learning, data mining, network science, biology, sociology, and more, and contains hundreds of link prediction algorithms now. Most algorithms used in scientific applications in sociology or biology are unsupervised ``topological'' predictors, which are simple or complicated functions of the local network structure around a pair of unconnected nodes $i,j$, e.g., their degrees or number of common neighbors~\cite{lu2011link}. More sophisticated techniques can define a probabilistic or Bayesian generative model of the entire network~\cite{clauset2009hierarchical}, as in the popular stochastic block model~\cite{holland1983stochastic}. The current state-of-the-art for link prediction is either meta-learning methods like model stacking~\cite{ghasemian2020stacking,he2024temporal,vankleunen2024traits} or various graph neural network (GNN) methods~\cite{kipf2017semisupervised,hamilton2018inductive}, although no studies so far have compared these two techniques directly. Similarly, LP studies often use different network benchmarks, different accuracy metrics, and different algorithms, making it difficult to assess the relative merits or performance of different methods.

As a meta-learning method, model stacking is particularly attractive for LP, as it can learn to combine any other predictors~\cite{ghasemian2020stacking}, both supervised and unsupervised, topological or model-based, etc. Past work has only evaluated random forests and gradient boosting~\cite{ghasemian2020stacking} as the supervised meta-learner. Model stacking has also been used to predict links in evolving networks~\cite{he2024temporal}, and with non-standard missing edge functions~\cite{he2024sampling}. It remains unknown whether other supervised techniques, such as logistic regression and support vector machines, may produce more accurate predictions when used with model stacking. Embedding techniques like graph neural networks are attractive for LP because of their flexibility, and ability to naturally exploit node attributes to improve predictions~\cite{zhang2018link,zhang2019inductive}. However, it remains unknown how well these methods perform in hard LP settings, e.g., without node attributes, or how they compare to model stacking methods. Furthermore, the LP benchmarks for GNNs are often small or structurally non-diverse, and it is unknown how their performance varies across structurally diverse networks, e.g., social vs.\ biological vs.\ economic networks.

\subsection{Our Contributions}
In this paper, we systematically investigate and compare the performance of six state-of-the-art LP algorithms on two hardware configurations using a large, structurally diverse benchmark of 550 real-world networks drawn from six scientific domains~\cite{ghasemian2019evaluating}. This corpus is composed of 22\% social, 21\% economic, 33\% biological, 12\% technological, 4\% information, and 7\% transportation networks (Fig.~\ref{fig:corpus}), and its networks span three orders of magnitude in scale, providing clear insights for how algorithms are likely to perform on very large networks. None of these networks include node or edge attributes, making the task of predicting links a structure-only prediction, the hardest LP setting. Further details of the benchmark are given in Appendix~\ref{appendix:corpus}.

We measure performance using two standard metrics of accuracy:\ the AUC, which quantifies a method's ability to distinguish missing links from non-missing links, and Top-k, which quantifies a method's ability to identify many missing links under a fixed prediction budget. We compare performance in terms of which algorithm is best, which algorithm is ``nearly-best,'' their mean performance, in general and by scientific domain, and we evaluate which scientific domains are the easiest and hardest for LP.

We then develop a model to predict algorithm performance using only measurable characteristics of the input network, e.g., transitivity, mean degree, degree assortativity, etc., and quantify this model's performance loss relative to an oracle that knows the best algorithm ahead of time. Finally, we provide general guidance for practitioners wanting to use these algorithms in practical applications, and provide a Python package that implements these algorithms.

% ----- FIGURE 1: network data sets --------
\begin{figure}[t!]
\centering
\includegraphics[width=0.98\columnwidth]{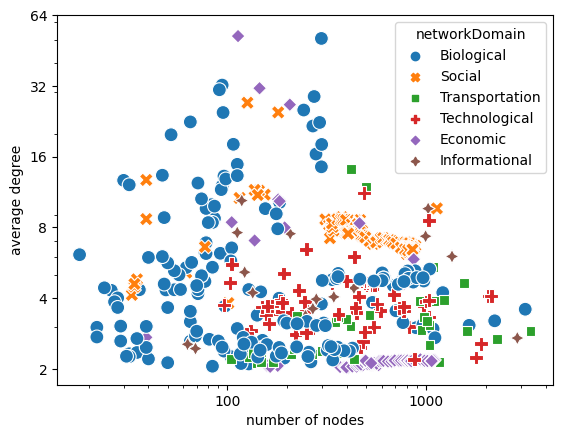}
\label{fig:corpus}
\vspace{-3mm}
\caption{Link prediction benchmark corpus, showing the average degree vs.\ number of nodes for 550 structurally diverse networks spanning six scientific domains.}
\label{fig:corpus}
\end{figure}
% ----------------------

\section{Materials and Methods}
This section describes (1)~the formal setup of the LP problem, (2)~the two classes of state-of-the-art algorithms and the six specific instances we evaluate here, (3)~the experimental design for algorithm training and hyperparmeter tuning using the empirical benchmark, and (4)~the two use-case-oriented evaluation metrics we use.

\vspace{3mm}\noindent\textbf{Link prediction is binary classification.}
Formally, the LP problem is a type of binary classification with highly unbalanced classes. Let $G=(V,E)$ denote a fully observed network with $V$~nodes and $E$~pairwise connections, with sizes~$n$ and~$m$, respectively. A subset of edges $E' \subset E$ are observed, chosen by a `missingness' function $E'=g(E)$ that defines an \textit{observed network} $G'=(V,E')$. In the hardest LP setting, where $G$ has no node or link attributes, the objective is to predict, solely from the observed edges $E'$, which pairs of unconnected nodes in the set $X = V \times V - E'$ are in fact missing links in the set $Y = E - E'$. In a sparse network, there will be $|Y|=O(n)$ true positives and $|X-Y|=\Theta(n^2)$ true negatives.

The LP literature typically simulates a missingness function~$g$ by retaining edges uniformly at random, with rate $\alpha$. Retaining edges in non-uniform ways is an interesting direction for future work~\cite{he2024sampling}. Here, we consider only the standard choice. In all experiments, we set $\alpha = 0.8$ so that 80\% of the ground truth edges are observed in~$E'$ and the remaining 20\% are the missing or held-out set~$Y$.

\vspace{3mm}\noindent\textbf{State-of-the-art algorithms.}
We evaluate and compare two classes of state-of-the-art LP algorithms---model stacking and graph neural networks---which have previously been shown separately to produce superior performance predicting missing links in real-world networks~\cite{ghasemian2020stacking,he2024temporal,vankleunen2024traits,kipf2017semisupervised,hamilton2018inductive,zhang2018link}. 

\textit{Model stacking} (also called stacked generalization) is an ensemble technique that learns to combine weakly predictive base models to construct an optimal predictive distribution~\cite{WOLPERT1992241,yao2018stackingbayes}. Stacking begins by training a set of $N$~base learners or level-0 models $\{f_1, f_2, \ldots, f_N\}$ on a common data set~$X$, and then ``stacking'' a meta-learner or level-1 model on top to learn to synthesize their outputs into a final prediction~$Y$~\cite{WOLPERT1992241}. Level-0 models can be any supervised or unsupervised technique. Each base learner $f_i$ maps an input $x \in X$ to a predicted output $f_i(x) \in \mathbb{R}$, and subsequently, the predictions $f_i(x)$ obtained on the training set are treated as additional training data for the meta-learner model $F$. Hence, the level-1 model learns to predict the true target $y\in \mathbb{R}$ by combining various level-0 model predictions~\cite{Breiman1996StackedRegressions}:\ 
%
%\begin{align}
$y_{{\rm final}} = F(f_1(x), f_2(x), \ldots, f_N(x))$. %\nonumber \enspace .
%\end{align}
For scientific applications where interpretability is important, users can select a level-1 model with good interpretability. Here, we evaluate four supervised learning algorithms for model stacking:\ random forests (RF)~\cite{Breiman2001}, extreme gradient boosting (XGB)~\cite{Chen2016}, logistic regression (LR)~\cite{Hosmer2013}, and support vector machines (SVM)~\cite{Cortes1995}. LR and SVM stacking for the LP problem are studied here for the first time. Background and technical details for these algorithms is provided in Appendix~\ref{appendix:level1}.

For level-0 models, we use 32 unsupervised and computationally lightweight topological functions of the input network, which have been used in prior LP studies, to derive 42 level-0 models. A complete list of these functions is given in Appendix~\ref{appendix:level0}. At a high level, these functions come in three flavors:\ global, node-level, and pair-level.
The 8 global predictors are network-level statistics like the number of nodes $n$, number of edges $m$, etc.\ that provide contextual information about a network's overall structure. The 10 node-level predictors are properties of individual nodes, e.g., degree $k_i$, etc., that provide localized context within a network, and we include these for both nodes in an unconnected pair $i,j\in X$. The 14 pairwise predictors quantify the structural relationship of the pair $i,j$, e.g., the number of common neighbors CN$(i,j)$, etc., directly informing the likelihood of a connection.

\textit{Graph neural network} (GNN) techniques are a broad class of algorithms that learn a neural embedding of a graph and missing links are predicted by the embedded proximity of unconnected pairs $i,j$. 
Within this class, we evaluate graph convolutional networks (GCN)~\cite{kipf2017semisupervised} and GraphSAGE (SAGE)~\cite{hamilton2018inductive}. 
GCN applies convolutional operations directly on the graph, such that each node can aggregate features from its neighbors, thereby capturing local patterns in graph connectivity in a rich node embedding. SAGE extends GCN by aggregating features from a sample of a node's neighbors rather than from all neighbors, which reduces its computational complexity and improves scalability to larger graphs. SAGE provides greater flexibility in the aggregation function, e.g., mean pooling, max pooling, etc., compared to GCN's sum-based function. This flexibility allows SAGE to adapt its node embeddings to a wider variety of graph structures and feature distributions~\cite{hamilton2018inductive}.

In the hard LP setting we consider here, where graphs have no node or link attributes, GCN and SAGE models are trained without predefined node features. Our own experiments with these methods indicate that using an identity matrix as the node feature input, i.e., allowing the model to learn a node feature matrix, yields superior performance. The learned embeddings are then combined with multi-layer perceptron (MLP) layers and we apply a sigmoid loss function to predict missing links in the network.

\vspace{3mm}\noindent\textbf{Experimental design.} 
To systematically evaluate the performance of these state-of-the-art LP algorithms, we compare four types of model stacking and two types of graph neural networks applied to all 550 networks in the benchmark, for two distinct performance metrics. Because GNN techniques can benefit from running on GPU hardware, for timing experiments, we evaluate both CPU and GPU implementations of the GNN algorithms.

To answer our research questions about which algorithms perform best, on which networks, and how well can such performance be predicted from the characteristics of the input, we developed a standardized training/testing/validation approach and applied it uniformly across all six algorithms, as follows.

First, we split each network~$G$ in the benchmark into `observed' and validation data sets by retaining an~$\alpha$ fraction of ground truth edges uniformly at random; the retained edges define an observed network $G'$ and the held out edges $Y=E-E'$ (true positives, or missing links) define the validation set for $G$. We note, however, random sampling can alter a network's topology, e.g., a triangle is only preserved if all three of its edges are retained by chance. To capture this source of variability during model training, we repeat the above `observation' procedure 10 times for each ground truth network $G$, producing a set of observed networks $\{G'_1,G'_2,\dots,G'_{10}\}$, and hence $5500$ total data points across all networks, each representing a randomly observed network topology.
When we evaluate the performance of a LP algorithm on validation data, we calculate any required features, e.g., the topological predictors, on the observed network structure $G'$ (not on the ground truth network $G$).

Second, we further split each observed network $G'$ into training and test data. For model stacking, this second split supports the supervised training of the level-1 meta-learner model, while for graph neural networks, the second split is used for hyperparameter tuning. Specifically, for each observed network $G'$ we retain an $\alpha$ fraction of observed edges uniformly at random; the retained edges define a training network $G''$ and the omitted edges define the positive examples of the test set $Y'=E'-E''$. Following best practices in the LP literature~\cite{ghasemian2020stacking}, we treat $X'=V\times V-E'$ as negative examples during training, and we compute all level-0 predictors in the stacked models from the training networks $G''$. For large networks, e.g., those with more than 10,000 non-edges, we uniformly subsample the non-edge set to 10,000 to form the negative class, and uniformly upsample the positive examples to the same amount before model training. This step balances the training classes and has been shown previously to produce better results in the LP problem~\cite{ghasemian2020stacking}.

The hyperparameter tuning for both model stacking algorithms (RF, XGB, LR, SVM) and graph neural network algorithms (GCN, SAGE) is done via grid search using $G''$, with model stacking using 5-fold cross-validation and graph neural networks using simple cross-validation (due to computational costs), to systematically explore a range of hyperparameter values~\cite{bergstra2012random}, ensuring that the tuned hyperparameters generalize well to unseen data and across different datasets. Parameters are optimized using the AUC metric, and once trained, we use the test data to evaluate whether the models can accurately distinguish unseen edges from non-edges. Technical details of the hyperparameter tuning approach for each algorithm are given in Appendix~\ref{appendix:hyper}.

\vspace{3mm}\noindent\textbf{Evaluation metrics.}
We evaluate and compare LP algorithm performance under two use cases. First, how well can an algorithm learn to distinguish true positives (missing links $Y$) from true negatives (non-edges $X-Y$)? The AUC is a threshold invariant metric that measures precisely this distinguishability, and is widely used in the LP literature, facilitating comparisons with past work~\cite{Yang_2014}. Second, how well can an algorithm maximize true positives under a fixed prediction budget? For example, an experimentalist who can conduct exactly $k$~measurements and wishes to recover as many true positives as possible. Top-k is a threshold-specific metric that measures precisely this utility~\cite{Topk}. We note that Top-k is closely related to other measures used in the LP literature, such as Average Precision~\cite{zhu2004recall} (also threshold invariant) and Hits@K~\cite{mohamed2020popularity}. However, in our experiments, these alternative measures provide little insight beyond what AUC and Top-k provide. Finally, we note that for small networks, if the number of edges to predict is less than 100, we set $k=10$; otherwise, we set $k=100$ regardless of network size.

\section*{Results}

\subsection*{Link prediction training times}

Training time is an important characteristic of LP algorithms, as the longer the training time, the more expensive any link predictions are to obtain, particularly when applied to very large systems. Applied to the 550 benchmark networks, we tracked the training times for all four stacked models and four versions of the GNN models (using CPU and GPU hardware). This experiment provides a broad benchmark to compare LP algorithm training times across all different types of networks. For model stacking approaches, the reported training times include the computational overhead of extracting topological features from the network structure, which is a prerequisite step before any machine learning model can be trained. This feature extraction phase often represents a significant portion of the total training time, particularly for networks with complex topological properties. 

\begin{figure}[t!]
\centering
\includegraphics[width=\columnwidth]{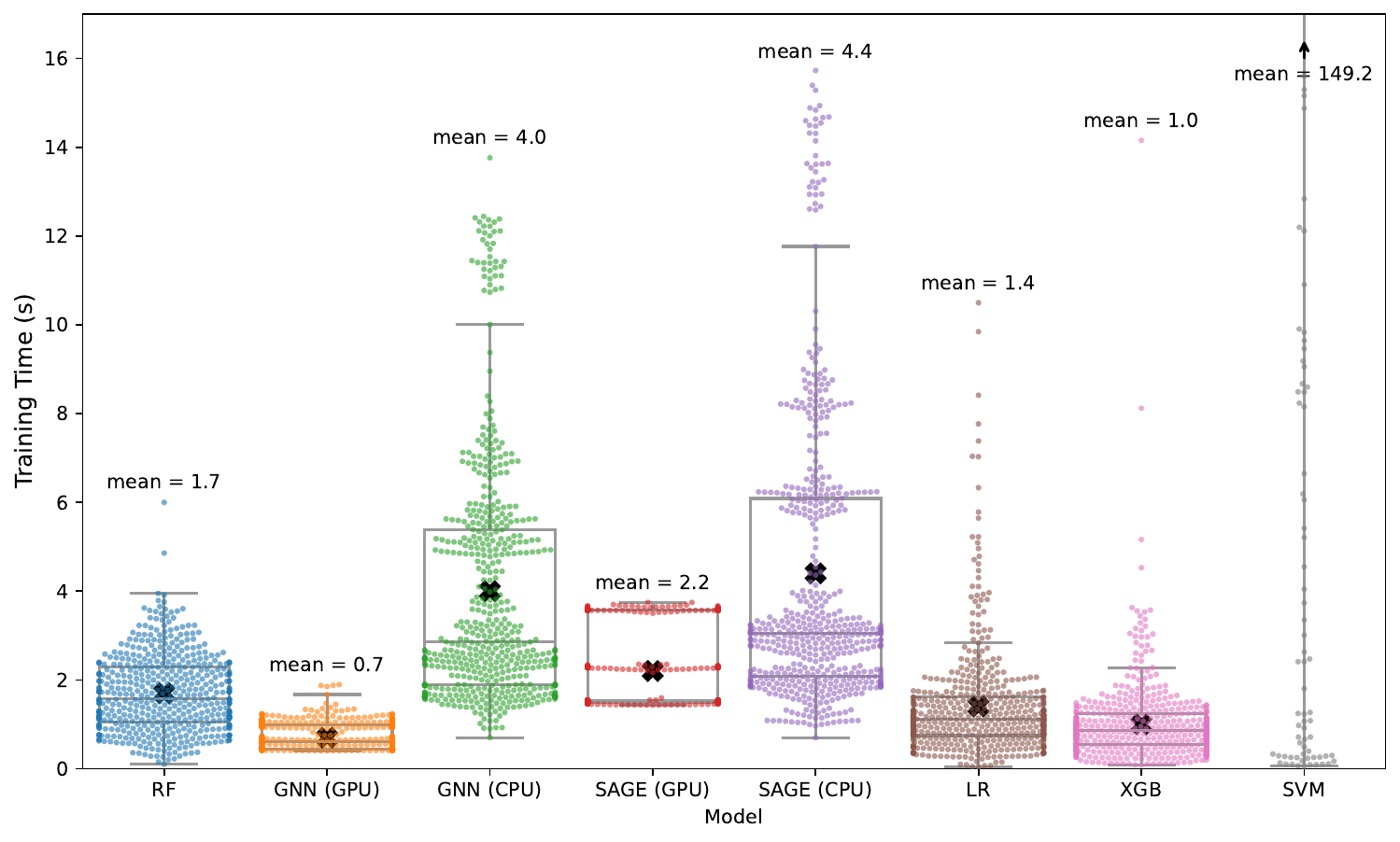}
\vspace{-5mm}
\caption{LP algorithm training times for six different link prediction models on 550 networks (see text for abbreviations), and two types of hardware. Means are marked in black, boxes show the median and 5-95\% quantiles.} 
\label{fig:time_box}
\end{figure}

Figure~\ref{fig:time_box} shows a swarm plot (with boxplot overlayed) of the resulting training times for each model, with mean values highlighted. Overall, we find that algorithms can be grouped into three categories. Model stacking (RF, LR, XGB) dominates the fast-training group of algorithms, which also includes GCN (GPU); these algorithms have mean training times of 0.7--1.7 seconds per network. The remaining GNN algorithms make up a slower-training group of algorithms, which take around 2-4x more time to train. Notably, the SAGE GPU implementation is in this group and is substantially slower than CPU implementations of RF, LR, and XBG for model stacking. Finally, SVM with model stacking is about 140 times slower to train than the fast-training algorithms. Hence, model stacking on commodity hardware is generally, but not always, substantially more efficient to train than GNNs for link prediction, even on specialized hardware.

%% --------- FIGURE 2: training times --------
%% -----------------------------------------

%% --------- FIGURE 3: best algorithm heat maps ----
\begin{figure*}[t!]
\centering
\begin{tabular}{cc}
\includegraphics[width=0.99\textwidth]{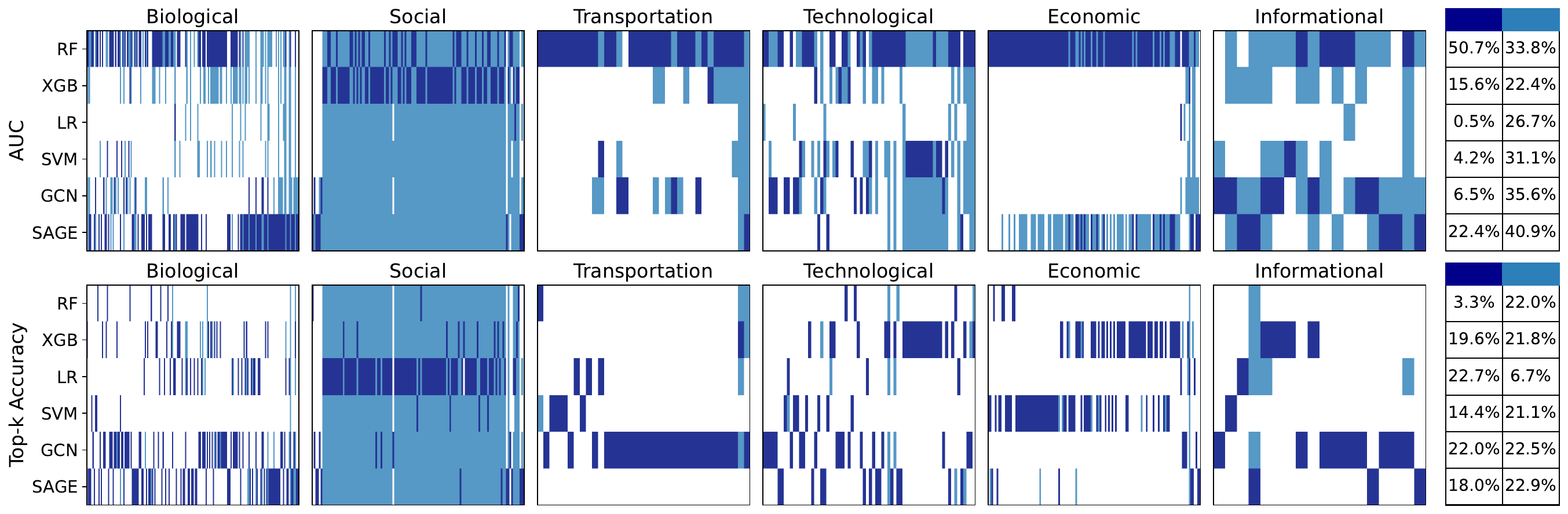}
\vspace{-2mm}
\end{tabular}
\caption{Best algorithm results for six state-of-the-art link prediction algorithms (see text for abbreviations) applied to 550 networks across six scientific domains, for two accuracy metrics (AUC and Top-k). For each network, the single algorithm yielding the highest mean accuracy is marked dark blue (``best''), and all algorithms with accuracy within 3\% of the highest accuracy are marked in light blue (``nearly-best''). Within a domain, networks are sorted in increasing order of their mean degree. Overall performance rates (best or nearly-best) by algorithm are given in the right-hand tables. Under AUC, RF stacking achieves best overall performance, followed by SAGE. Under Top-k, XGB stacking, GCN, and SAGE achieve overall good performance. All algorithms perform well on social networks, and all algorithms are best on some subset of networks.}
\label{fig:heatmap}
\end{figure*}
%% -----------------------------------------

\subsection*{Benchmarking link prediction performance}

% First Result
We first answer the question of whether one state-of-the-art LP algorithm exhibits superior performance in link prediction across all input networks. Additionally, we measure how competitive alternative algorithms are with the best algorithm, and whether and how algorithm performance varies systematically with the scientific domain of the network, e.g., are missing links easier to recover in social networks than biological networks? All results are means over 100 iterations per single validation split $G'$. We report results for both AUC (distinguishability of missing links) and Top-k (predictions on a budget) for applying all six algorithms to all 550 benchmark networks.

\vspace{3mm}\noindent\textbf{Which algorithm is best?}
Separately for AUC and for Top-k, Figure~\ref{fig:heatmap} indicates the best-performing algorithm (highest mean accuracy) for each of the 550 benchmark networks, grouped by scientific domain, and all ``nearly-best'' algorithms (mean accuracy within a 3\% margin of the highest accuracy on that same network). 

Under AUC, RF model stacking achieves the best overall accuracy across networks:\ it is best for 50.7\% of input networks and nearly-best for an additional 33.8\% of networks (a total of 84.5\%). SAGE is best for 22.4\% of networks and nearly-best for 40.9\% (63.3\% total). Other algorithms are generally less accurate, although every algorithm is best on some networks (even SVM model stacking).

Under Top-k, LR model stacking has the highest rate of best accuracy (22.7\%), closely followed by GCN (22.0\%), although GCN is nearly-best far more often (22.5\% vs.\ 6.7\%). Both XGB model stacking (19.6\% best, 21.8\% nearly-best) and SAGE (18.0\% best, 22.9\% nearly-best) yield similar accuracies to GCN, while RF model stacking has the lowest best accuracy rate (3.3\%). Again, however, every algorithm is best on some networks.

\begin{figure*}[t!]
\centering
\begin{tabular}{cc}
\includegraphics[width=0.98\textwidth]{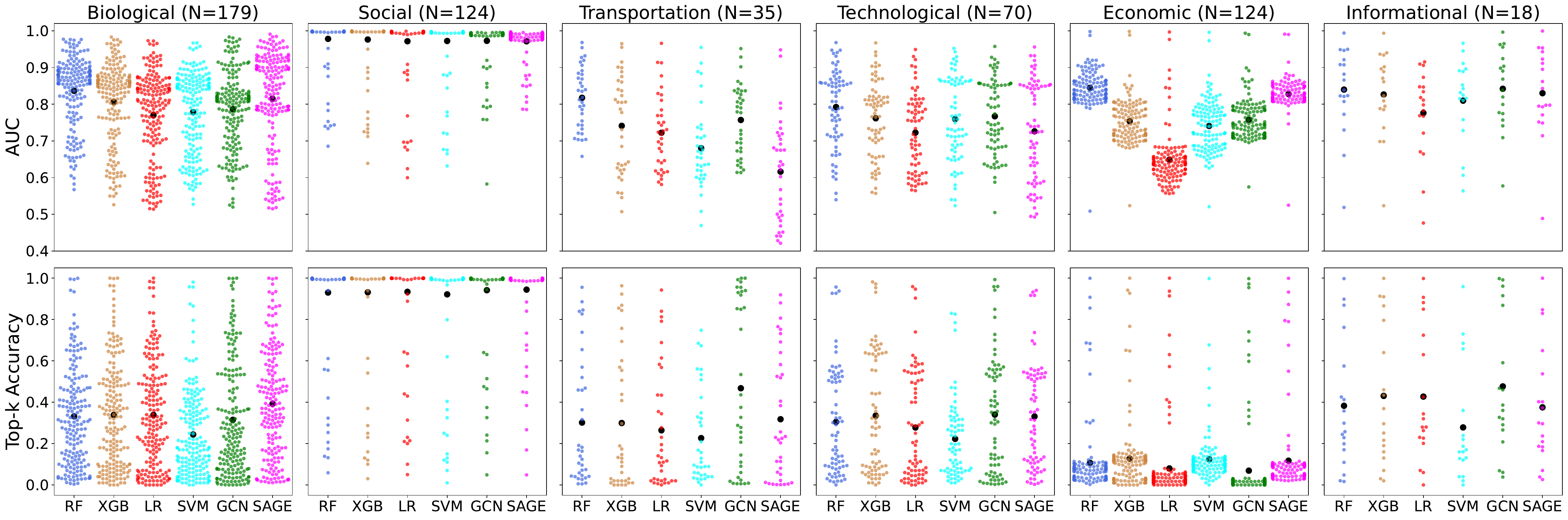}
\vspace{-3mm}
\end{tabular}
\caption{Distributions of algorithm accuracies on the Link Prediction problem for six algorithms (see text) applied to each of 550 benchmark networks, divided into six scientific domains:\ Biological ($N=179$), Social ($N=124$), Transportation ($N=35$), Technological ($N=70$), Economic ($N=124$), and Informational ($N=18$). Top row shows results for AUC (distinguishing missing links from non-edges); bottom row shows results for Top-k accuracy (accuracy within a prediction budget). Black dots mark the mean performance for a given algorithm within the domain. Across domains, all algorithms perform well on most social networks under AUC and Top-k, and exhibit more variable performance otherwise.}
\label{fig:swarm}
\end{figure*}

Figure~\ref{fig:swarm} elaborates these results by showing the mean accuracy of every algorithm on every benchmark network in a set of swarm plots, one for each domain. In every domain, we find a broad distribution of accuracies for each algorithm, indicating that variations in network structure strongly influence accuracy on the LP problem. At the same time, across benchmark networks, RF stacking yields the highest mean AUC $=0.86\pm0.10$ by some margin (Table~\ref{tab:auc_metrics}), and SAGE yields the highest mean Top-k$=0.44\pm0.37$, closely followed by XGB stacking and GCN (Table~\ref{tab:topk_metrics}).

Together, this systematic application of state-of-the-art LP algorithms on a large and structurally diverse benchmark shows that no algorithm is best over all networks, regardless of whether we prioritize missing link distinguishability (AUC) or accurate predictions on a budget (Top-k), and algorithm accuracy can vary broadly. Moreover, both model stacking and graph neural network algorithms achieve very strong overall accuracies:\ RF and SAGE under AUC, and XGB, GCN, and SAGE under Top-k.

\vspace{3mm}\noindent\textbf{Performance depends on domain.}
These results also reveal how LP accuracy varies across scientific domains, which answers the question of whether the LP problem can be easier for some types of networks, while being harder for others.

% AUC
%   domain  count_second_best_auc  count_nets  Sj/5*Nj 
% Biological     184         179  0.2056
% Economic       106         124  0.1710
% Informational   42          18  0.4667
% Social         573         124  0.9242
% Technological  114          70  0.3257
% Transportation  29          35  0.1657

% Topk
%    domain  count_second_best_top  count_nets  Sj/5*Nj 
% Biological      44         179  0.0492
% Economic        19         124  0.0306
% Informational    6          18  0.0667
% Social         555         124  0.8952
% Technological   14          70  0.0400
% Transportation   6          35  0.0343

From the perspective of the best and nearly-best algorithms, Figure~\ref{fig:heatmap} reveals a stark difference:\ in 92.4\% of cases for social networks under AUC and 89.5\% under Top-k, an algorithm achieves accuracy within 3\% of the most accurate algorithm. No other domain comes close to this level of general accuracy across algorithms. Moreover, the mean accuracies on most social networks are very close to their maximal value (Fig.~\ref{fig:swarm}):\ mean AUC ranges from 0.97--0.98, and mean Top-k ranges from 0.92--0.94 (Tables~\ref{tab:auc_metrics} and~\ref{tab:topk_metrics}).

In contrast, in only 17\% of cases for transportation and economic networks under AUC and only 3\% under Top-k, an algorithm achieves accuracy within 3\% of the best. Additionally, mean accuracies are substantially lower (Fig.~\ref{fig:swarm}):\ mean AUC ranges from 0.72--0.82 for transportation and from 0.65--0.84 for economic networks, while mean Top-k ranges from 0.23--0.47 for transportation and from 0.07--0.13 for economic networks (Tables~\ref{tab:auc_metrics} and~\ref{tab:topk_metrics}). Together, these results indicate that social networks are the easiest setting for predicting missing links, while transportation and economic networks are the hardest, and biological networks in between.

Under AUC, across these domains, the algorithm that yields the highest mean accuracy varies considerably (Table~\ref{tab:auc_metrics}). On biological networks, RF stacking and SAGE have highest mean accuracies, with SAGE excelling on higher-degree networks. On technological and informational networks, no single algorithm dominates, while on economic networks, RF stacking and SAGE consistently excel, while on transportation networks, RF stacking is typically most accurate, although GCN also does well.

Under Top-k, accuracies also vary considerably by domain (Table~\ref{tab:topk_metrics}). On biological networks, GCN and SAGE perform competitively. On technological and informational networks, we again find high variability, with all algorithms except SVM performing similarly. On economic networks, XGB and SVM stacking tend to be most accurate, contrasting with the pattern under AUC, and on transportation networks, GCN is most accurate.

% FIGURE 4

%%%%%% TABLE 1 : AUC accuracy by domain
\begin{table}[t!]
\setlength{\tabcolsep}{3pt} % Adjust column spacing for clarity
\renewcommand{\arraystretch}{1.2} % Reduce spacing between rows globally
\centering
{\small % Start larger text size
\begin{tabular}{lccccccc}
\hline
\textbf{Model} & \textbf{All} & \textbf{Bio.} & \textbf{Social} & \textbf{Trans.} & \textbf{Tech.} & \textbf{Econ.} & \textbf{Info.} \\
\hline
RF   & \textbf{0.86} & \textbf{0.84} & \textbf{0.98} & \textbf{0.82} & \textbf{0.79} & \textbf{0.84} & \textbf{0.84} \\[-0.5em]
     & (0.10) & (0.09) & (0.06) & (0.08) & (0.10) & (0.05) & (0.12) \\[-0.1em] % Add space after std row
XGB  & 0.82 & 0.81 & \textbf{0.98} & 0.76 & 0.74 & 0.75 & \textbf{0.83} \\[-0.5em]
     & (0.13) & (0.11) & (0.07) & (0.10) & (0.13) & (0.06) & (0.11) \\[-0.1em] % Add space after std row
LR   & 0.78 & 0.77 & 0.97 & 0.72 & 0.72 & 0.65 & 0.78 \\[-0.5em]
     & (0.15) & (0.12) & (0.08) & (0.11) & (0.10) & (0.07) & (0.12) \\[-0.1em] % Add space after std row
SVM  & 0.81 & 0.78 & 0.97 & 0.76 & \textbf{0.76} & 0.74 & 0.81 \\[-0.5em]
     & (0.13) & (0.11) & (0.08) & (0.11) & (0.11) & (0.07) & (0.11) \\[-0.1em] % Add space after std row
GCN  & 0.82 & 0.79 & 0.97 & 0.76 & \textbf{0.77} & 0.76 & \textbf{0.84} \\[-0.5em]
     & (0.12) & (0.10) & (0.06) & (0.10) & (0.10) & (0.10) & (0.11) \\[-0.1em] % Add space after std row
SAGE & 0.83 & \textbf{0.82} & 0.97 & 0.73 & 0.62 & \textbf{0.83} & \textbf{0.83} \\[-0.5em]
     & (0.14) & (0.13) & (0.04) & (0.14) & (0.14) & (0.04) & (0.12) \\[-0.1em] % Add space after std row
\hline
\end{tabular}
} % End larger text size
\caption{AUC accuracy (mean $\pm$ std) on link prediction for each model by scientific domain; scores in a given column that are statistically indistinguishable (t-test) are boldfaced.}
\label{tab:auc_metrics}
\vspace{-3em}
\end{table}
%%%%%%%%%%%%%%%%%%%%%%%%%%%%%%%%

%%%%%% TABLE 2 : Top-k accuracy by domain
\begin{table}[t!]
\setlength{\tabcolsep}{3pt} % Minimize column spacing
\renewcommand{\arraystretch}{1.2} % Reduce spacing between rows globally
\centering
{\small % Start larger text size
\begin{tabular}{lccccccc}
\hline
\textbf{Model} & \textbf{All} & \textbf{Bio.} & \textbf{Social} & \textbf{Trans.} & \textbf{Tech.} & \textbf{Econ.} & \textbf{Info.} \\
\hline
RF   & \hspace{0.1em} \textbf{0.41} & \hspace{0.1em}0.33 & \hspace{0.1em}\textbf{0.93} & \hspace{0.1em}0.30 & \hspace{0.1em}0.30 & \hspace{0.1em}0.11 & \hspace{0.1em} 0.38\\[-0.5em]
     & (0.37)             & (0.24)             & (0.21)             & (0.31)             & (0.26)             & (0.17)             & (0.31)             \\[-0.1em] % Add space after std row
XGB  & \hspace{0.1em} \textbf{0.42} & \hspace{0.1em}0.34 & \hspace{0.1em}\textbf{0.93} & \hspace{0.1em}0.30 & \hspace{0.1em}\textbf{0.34} & \hspace{0.1em}\textbf{0.13} & \hspace{0.1em}\textbf{0.43} \\[-0.5em]
     & (0.38)             & (0.26)             & (0.22)             & (0.33)             & (0.29)             & (0.18)             & (0.34)             \\[-0.1em] % Add space after std row
LR   & \hspace{0.1em} \textbf{0.40} & \hspace{0.1em}0.34 & \hspace{0.1em}\textbf{0.93} & \hspace{0.1em}0.26 & \hspace{0.1em}0.28 & \hspace{0.1em}0.08 & \hspace{0.1em}\textbf{0.43} \\[-0.5em]
     & (0.38)             & (0.25)             & (0.21)             & (0.29)             & (0.26)             & (0.18)             & (0.32)             \\[-0.1em] % Add space after std row
SVM  & \hspace{0.1em}0.37 & \hspace{0.1em}0.24 & \hspace{0.1em} 0.92 & \hspace{0.1em}0.23 & \hspace{0.1em}0.22 & \hspace{0.1em}\textbf{0.12} & \hspace{0.1em}0.28 \\[-0.5em]
     & (0.36)             & (0.21)             & (0.23)             & (0.23)             & (0.19)             & (0.12)             & (0.28)             \\[-0.1em] % Add space after std row
GCN  & \hspace{0.1em} \textbf{0.42} & \hspace{0.1em}0.31 & \hspace{0.1em}\textbf{0.94} & \hspace{0.1em}\textbf{0.47} & \hspace{0.1em}\textbf{0.34} & \hspace{0.1em}0.07 & \hspace{0.1em}\textbf{0.48} \\[-0.5em]
     & (0.40)             & (0.27)             & (0.19)             & (0.40)             & (0.28)             & (0.19)             & (0.33)             \\[-0.1em] % Add space after std row
SAGE & \hspace{0.1em} \textbf{0.44} & \hspace{0.1em}\textbf{0.39} & \hspace{0.1em}\textbf{0.94} & \hspace{0.1em}0.32 & \hspace{0.1em}\textbf{0.33} & \hspace{0.1em}\textbf{0.12} & \hspace{0.1em}0.38 \\[-0.5em]
     & (0.37)             & (0.26)             & (0.17)             & (0.30)             & (0.26)             & (0.18)             & (0.29)             \\[-0.1em] % Add space after std row
\hline
\end{tabular}
} % End larger text size
\caption{Top-k accuracy (mean $\pm$ std) on link prediction for each model by scientific domain; scores in a given column that are statistically indistinguishable (t-test) are boldfaced.}
\label{tab:topk_metrics}
\vspace{-3em}
\end{table}
%%%%%%%%%%%%%%%%%%%%%%%%%%%%%%%%

\section{Network structure predicts performance}
Network structure is known to vary systematically with scientific domain~\cite{ikehara2017structure}, e.g.,  social networks tend to have far more triangles than other types of networks. How well can these structural patterns predict LP algorithm performance?

\vspace{3mm}\noindent\textbf{Exploratory network analysis.}
We begin by characterizing how link prediction accuracy varies with specific network features like mean degree $\langle k \rangle$, degree assortativity~$r$ (the tendency for nodes of similar degree to be connected), mean local clustering $\langle C_i \rangle$ (the density of connections among a node's neighbors), and the mean geodesic path length $\langle \ell \rangle$. 

Figure~\ref{fig:Trend} in Appendix~\ref{appendix:Trend} plots AUC and Top-k for all six algorithms vs.\ these four network features. The results show that algorithm performance on both metrics tends to improve with increasing $\langle k \rangle$, more positive~$r$, and greater $\langle C_i \rangle$. In contrast, performance tends to improve with smaller $\langle \ell \rangle$, although the association is weak. These patterns are consistent with better link prediction performance occurring when nodes are better connected, when neighborhoods are relatively more dense, and when networks are more compact.

Under AUC, RF stacking tends to produce higher accuracies than other methods across the range of network topological patterns, with SAGE often being second, indicating that the performance of these methods is less sensitive to changes in network topology. Notably, RF stacking is generally more accurate than other algorithms on more sparse networks, i.e., $\langle k \rangle<4$, and on networks with degree disassortativity, i.e., $r<0$. Under Top-k, there is little difference in algorithm performance with $\langle k \rangle$ or $r$, while SAGE and GCN are often slightly more accurate across values of $\langle C_i \rangle$. Additional discussion of these results is provided in Appendix~\ref{appendix:Trend}.

\vspace{3mm}\noindent\textbf{Predicting performance.}
We now measure how well algorithm performance on the LP problem can, in fact, be predicted ahead of time from measurable characteristics of an input network. Operationalizing this analysis as a multivariate prediction problem allows us to account for the correlations among network features induced by the non-independence of edges, and to identify the most important network features that influence algorithm performance.

The feature set for this prediction problem includes the four variables from the exploratory network analysis along with two more variables:\ the variance of the degree distribution $\sigma_k^2$ and the network size $n$, which provide additional contextual information. For the same set of observed and validation splits used to measure mean algorithm performance in the previous sections, we associate the corresponding AUC and Top-k accuracies for each algorithm, and then calculate the corresponding network feature set for each of the observed networks $G'$ (each of ten random variants). We then divide these splits into an 80\% training set and 20\% test set at the level of networks, so that all topological variants of a single network appear together in either the training or the test subset.

To learn a model of algorithm accuracy, we train random forest regression models to predict AUC and Top-k link prediction accuracy solely from the network measures. We repeat the training process 100 times with different network-level training/test splits to capturing variability in the results. Additional technical details are given in Appendix~\ref{appendix:regmod}. Overall, the prediction models achieve excellent results, with average adjusted $r^2=0.988$ for AUC and $r^2=0.911$ for Top-k accuracy, indicating that these six features can explain much of the observed variance in link prediction accuracy and that prediction accuracy is highly predictable from a network's structure.

%%%% FIGURE 5 : gini importances, predicting AUC/Topk
\begin{figure}[t!]
\centering
\includegraphics[width=0.99\linewidth]{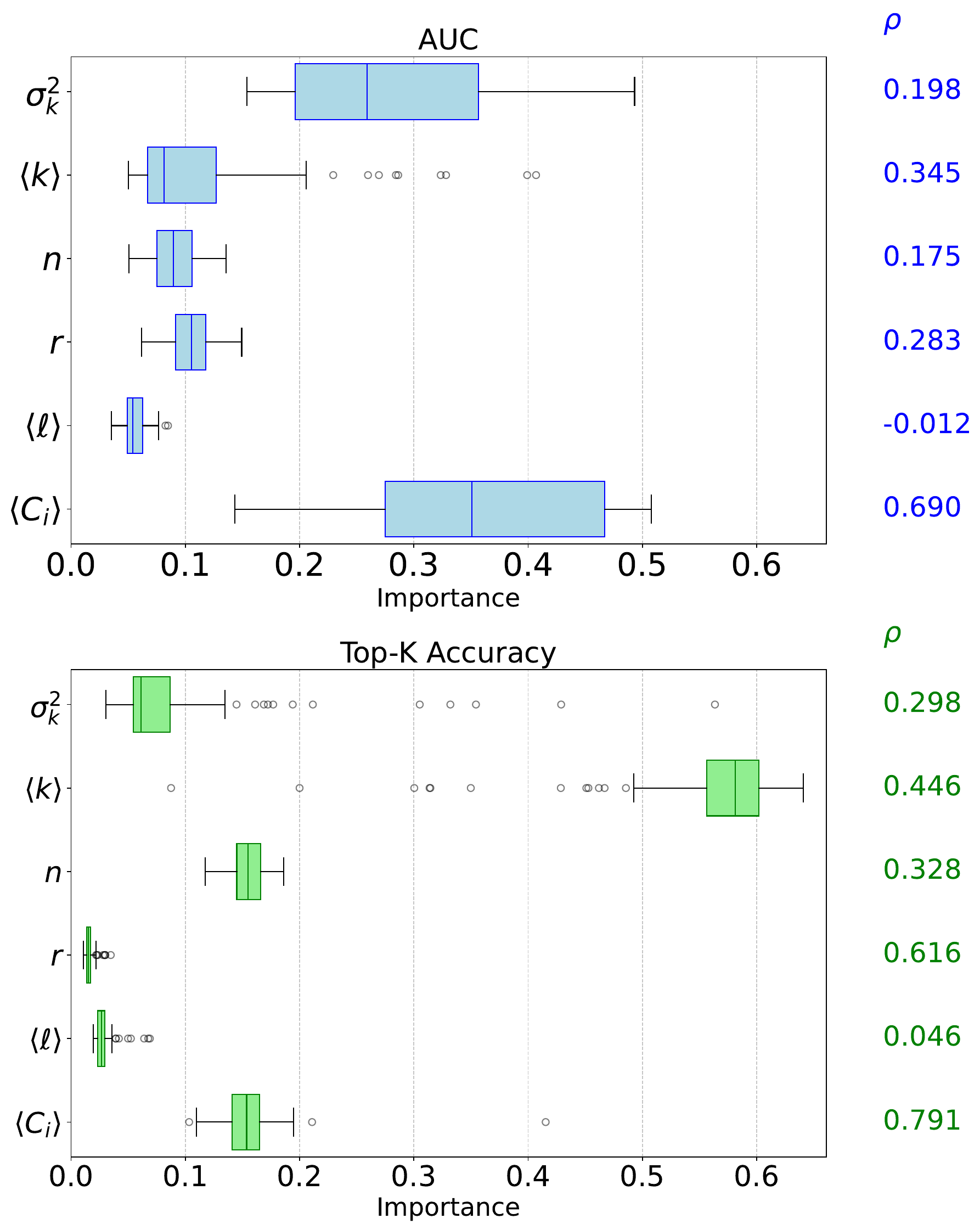}
\caption{Gini importance distributions for a random forest model trained to predict link prediction accuracy (AUC and Top-k) from six measurable characteristics of an input network (see text for definitions). Pearson correlations $\rho$ for network features and algorithm accuracy are given on the right. Networks with higher mean local clustering $\langle C_i \rangle$ and degree variance $\sigma_k^2$ tend to have easier to distinguish missing links vs.\ non-edges (AUC), while a higher mean degree $\langle k \rangle$ strongly improves predictions on a budget (Top-k).}
\label{fig:predictacc}
\vspace{-2em}
\end{figure}
%%%%%%%%%%%%%%%%%%%%%%%%%%%%%%%

Figure~\ref{fig:predictacc} shows the resulting Gini importances for topological features in the AUC and Top-k models. For AUC, mean local clustering $\langle C_i \rangle$, variance of degrees $\sigma_k^2$, and to a lesser extent mean degree $\langle k \rangle$ are most important. For Top-k accuracy, mean degree $\langle k \rangle$ is most important by a substantial margin, indicating that a higher mean degree increases the chance that true missing links rank among the top predictions. Mean local clustering $\langle C_i \rangle$ and network size $n$ are also important but much less so than mean degree.

Together, these results indicate that performance on the LP problem is highly predictable from a network's measurable characteristics, although which features are important depends on whether the prediction goal is distinguishing missing links from non-edges (AUC) or making predictions on a budget (Top-k). At the same time, these results indicate that the LP problem is generally harder on sparse networks (lower $\langle k \rangle$), which have fewer edges to learn from.

\section{Optimizing link prediction via meta-learning}
We can exploit the predictability of algorithm accuracy from network characteristics, and the fact that no algorithm is superior on all networks, by introducing a meta-learner algorithm that predicts which particular algorithm would yield the highest accuracy, based only on the input network's structural features. A researcher can then use this meta-learner to select the best LP algorithm for a particular network, without having to run all six state-of-the-art algorithms. We compare the meta-learner algorithm to an alternative that uses an oracle to make perfect predictions about which LP algorithm is best for a particular network, thus providing an upper bound on the meta-learner's possible performance.

We consider two versions of this meta-learner, a full version (\textit{Model~1}) that learns to choose among all six algorithms, and a restricted version (\textit{Model~2}) that learns to choose only between RF stacking and SAGE, the two algorithms with the best overall performances. We compare each meta-learner algorithm against a corresponding oracle algorithm that runs all candidate algorithms on the input network, measures the corresponding AUC or Top-k accuracy, and picks the top performer. Comparing meta-learner accuracy with that of the oracle quantifies the loss in accuracy due to learning from data to predict the best algorithm. For this task, we use the same feature and algorithm performance data as in the previous section, as well as the same network-level 80-20 train-test splits among networks in the corpus. Additional technical details are given in Appendix~\ref{appendix:meta}.

%%%%% FIGURE 6 : meta-learner performance
\begin{figure}[t!]
\centering
\includegraphics[width=0.98\columnwidth]{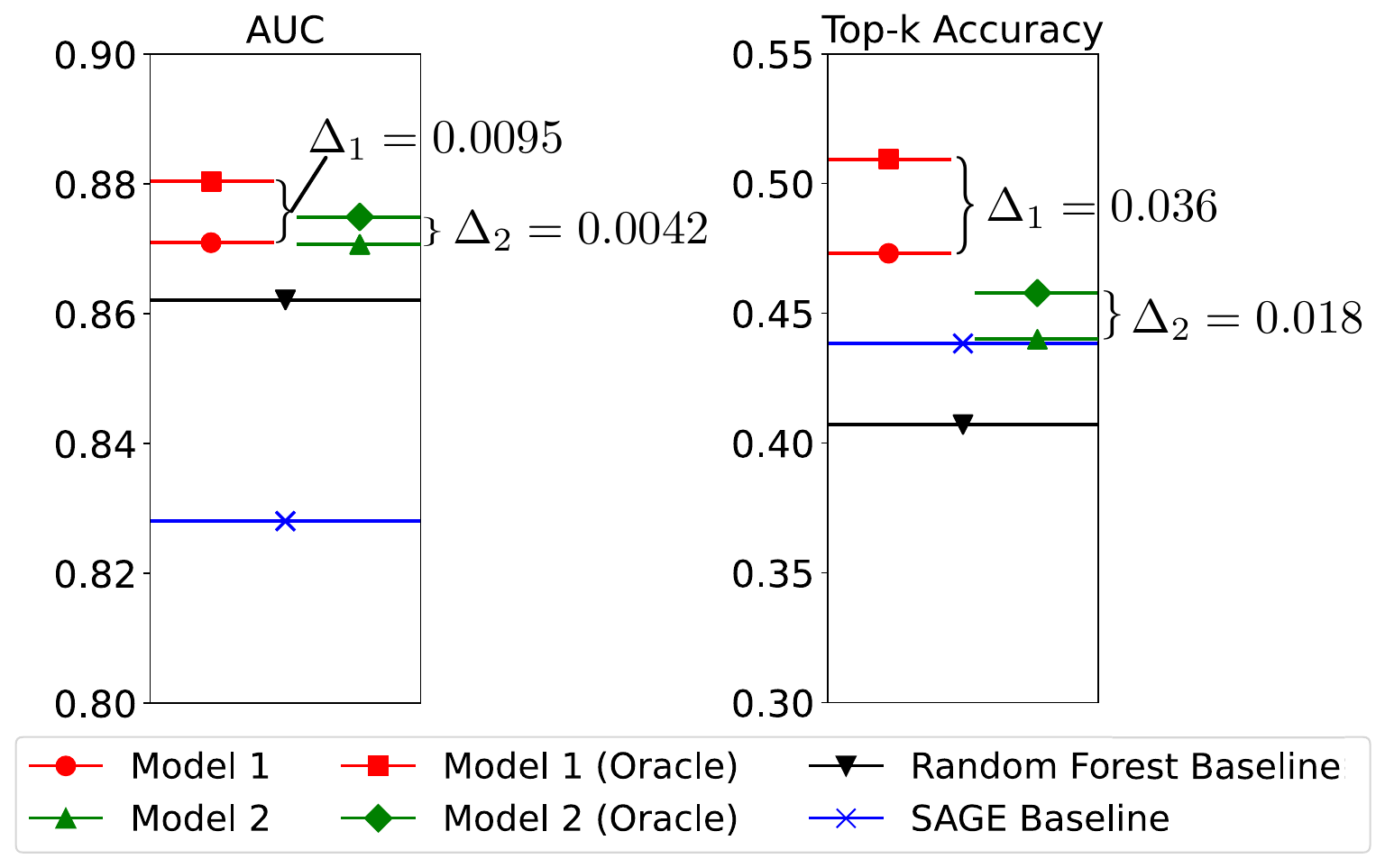}
\vspace{-3mm}
\caption{Mean link prediction performance of full (Model~1) and restricted (Model~2) meta-learner algorithms vs.\ oracle and baseline algorithms, showing substantial improvement in accuracy from meta-learning. Model~1 learns to choose among all six algorithms, while Model~2 learns to choose between RF stacking or SAGE.}
\label{fig:my_meta_learner}
%\vspace{-2em}
\end{figure}
%%%%%%%%%%%%%%%%%%%%%%%%%%%%%%%%

%                       Topk  Accuracy
% 0                  Model 1  0.476146
% 1                  Model 2  0.442285
% 2         Model 1 (Oracle)  0.511720
% 3         Model 2 (Oracle)  0.460430
% 4  Random Forrest Baseline  0.411506
% 5            SAGE Baseline  0.440643

%                        AUC  Accuracy 
% 0                  Model 1   0.872497
% 1                  Model 2   0.872455
% 2         Model 1 (Oracle)   0.881997
% 3         Model 2 (Oracle)   0.876698
% 4  Random Forrest Baseline   0.864707
% 5            SAGE Baseline   0.830591

%%%%%% TABLE: Meta-Learner AUC Results
\begin{table*}[t!]
\setlength{\tabcolsep}{5pt}
\renewcommand{\arraystretch}{1.1}
\centering
\caption{Meta-learner predicted and actual AUC accuracy and best model, for three representative large-scale networks. The true best AUC is marked in boldface.}
\label{tab:meta_auc}
{\small
\begin{tabular}{lrr|lll|rrrrr}
\hline
\textbf{Network} & \textbf{$n$} & \textbf{$m$} & \textbf{Pred. Model} & \textbf{Best Model} & \textbf{Pred. AUC} & \textbf{LR} & \textbf{RF} & \textbf{XGB} & \textbf{GCN} & \textbf{SAGE} \\
\hline
Facebook Page-Page~\cite{snap_facebook_pagepage} & 22,470 & 171,002 & RF   & XGB  & 0.9604 & 0.7950 & 0.9774 & \textbf{0.9832} & 0.9641 & 0.9381 \\
GitHub Social~\cite{snap_github_social}       & 37,700 & 289,003 & XGB  & XGB  & 0.9236 & 0.9208 & 0.9539 & \textbf{0.9591} & 0.9321 & 0.9416 \\
Deezer Europe User  ~\cite{snap_deezer_europe}     & 28,281 & 92,752  & RF   & XGB  & 0.9845 & 0.8359 & 0.9091 & \textbf{0.9240} & 0.8906 & 0.8609 \\
\hline
\end{tabular}
}
\end{table*}

%%%%%% TABLE: Meta-Learner Top-k Results
\begin{table*}[t!]
\setlength{\tabcolsep}{5pt}
\renewcommand{\arraystretch}{1.15}
\centering
\caption{Meta-learner predicted and actual Top-k accuracy and best model, for three representative large-scale networks. The true best Top-k is marked in boldface.}
\label{tab:meta_topk}
{\small
\begin{tabular}{lrr|lll|rrrrr}
\hline
\textbf{Network} & \textbf{$n$} & \textbf{$m$} & \textbf{Pred. Model} & \textbf{Best Model} & \textbf{Pred. Top-k} & \textbf{LR} & \textbf{RF} & \textbf{XGB} & \textbf{GCN} & \textbf{SAGE} \\
\hline
Facebook Page-Page ~\cite{snap_facebook_pagepage}& 22,470 & 171,002 & GCN  & RF   & 0.9466 & 0.9336 & \textbf{0.9998} & 0.9984 & 0.9931 & 0.9920 \\
GitHub Social~\cite{snap_github_social}       & 37,700 & 289,003 & LR   & SAGE & 0.9315 & 0.9971 & 0.9980 & 0.9988 & 0.9967 & \textbf{1.0000} \\
Deezer Europe  ~\cite{snap_deezer_europe}     & 28,281 & 92,752  & GCN  & GCN  & 0.8686 & 0.7449 & 0.9465 & 0.9560 & \textbf{1.0000} & \textbf{1.0000} \\
\hline
\end{tabular}
}
\end{table*}

Under both AUC and Top-k accuracy measures, the full meta-learner algorithm (Model~1) substantially outperforms both RF stacking and SAGE baselines across the 550 networks in the corpus (Fig.~\ref{fig:my_meta_learner}), achieving a mean $\textrm{AUC}=0.87$ and mean $\textrm{Top-k}=0.47$ (cf.\ with Tables~\ref{tab:auc_metrics} and~\ref{tab:topk_metrics}). This improved performance comes at a cost, taking 6.7-11x longer per network than SAGE, on average; however, the meta-learner is 42-52\% faster than SAGE if we exclude the cases where the full meta-learner selects SVM stacking (8-10\% of cases), 

The restricted meta-learner performs equally well over the baselines under AUC (mean $\textrm{AUC}=0.87$), and provides only a marginal improvement under Top-k (mean $\textrm{Top-k}=0.44$), while being on average 5-29\% faster than SAGE. In practice, full and restricted models often make the same prediction for AUC, as RF stacking is the most common `best' algorithm across networks (Fig.~\ref{fig:heatmap}). However, the two models produce more divergent predictions for Top-k accuracy, under which we found a wider range of best algorithms across networks (Fig.~\ref{fig:heatmap}) that is only accessible to the full model.

In both cases, learning to choose the best algorithm induces a modest performance cost, with the oracle algorithms outperforming their corresponding meta-learners:\ full model, $\Delta\textrm{AUC}=0.01$, mean $\Delta\textrm{Top-k}=0.04$; and restricted model, $\Delta\textrm{AUC}=0.004$, mean $\Delta\textrm{Top-k}=0.02$ (Fig.~\ref{fig:my_meta_learner}). These gaps indicate both room for improvement by the meta-learner with a better feature set and an upper limit to that improvement with the current set of LP algorithms.

\section{Meta-learner performance on large-scale networks}

To evaluate the practical utility of the meta-learner for scalable link prediction, we benchmark its performance on three large real-world networks from the SNAP repository:\ Facebook Page-Page~\cite{snap_facebook_pagepage}, GitHub Social~\cite{snap_github_social}, and Deezer Europe~\cite{snap_deezer_europe}. The scale and topological complexity of these networks represent a challenging case for traditional algorithm selection approaches.

For each network, we assess the meta-learner's ability to
\begin{enumerate}
\item predict which model is the best-performing model (under AUC or Top-k accuracy); and
\item estimate the expected accuracy (AUC or Top-k) of the best-performing model.
\end{enumerate}
We then compare these predictions against the actual performance of all candidate algorithms.

The results of this experiment are summarized in Tables~\ref{tab:meta_auc} and~\ref{tab:meta_topk}. All meta-learner predictions are based solely on structural properties of the input graph, i.e., the meta-learner's predictions are made without training any link-prediction algorithm itself. Bold entries indicate the true best-performing model under each metric.

Across all cases, the meta-learner consistently selects models whose performance closely approximates the choice that an oracle would make it if ran every link prediction algorithm first and selected the best-performing choice. Moreover, the performance gap between the meta-learning and such an oracle is typically quite low. This experiment demonstrates that the meta-learner offers a highly scalable and computationally efficient alternative to exhaustive benchmarking. This feature of the meta-learner enables rapid and accurate algorithm selection for large-scale networks where running multiple link prediction algorithms would be computationally expensive.

\section{Discussion and Conclusions}

Using a structurally diverse benchmark of 550 real-world networks, we systematically evaluate and compare the performance of different state-of-the-art approaches for the hard version of the Link Prediction (LP) problem where node attributes (node features) are not available and predictions must be made solely on the basis of observed links. These state-of-the-art techniques use either model stacking, a form of meta-learning, or graph neural networks, and we evaluate six specific algorithms within these two approaches.

We show that (1)~no LP algorithm is best on all networks, regardless of what accuracy measure is used; (2)~networks from some scientific domains are categorically easier for LP algorithms, with social networks being easiest by far (all algorithms perform very well), and economic or transportation networks being hardest (most algorithms perform poorly), 
with biological networks falling in between; and (3)~a network's structural patterns are highly predictive of LP algorithm accuracy, with variations in mean degree $\langle k \rangle$, local clustering $\langle C_i\rangle$, and degree variance $\sigma^2_k$ being most important. And, we show that these findings can be operationalized into a scalable meta-learning algorithm that can predict the accuracy of the best-performing LP algorithm on a particular network, using only the network's structure as input. We provide a Python package that implements this meta-learning LP algorithm that can be applied to any network to make accurate link predictions.

Across our experiments, when considering both accuracy and computational cost, we find that random forest (RF) stacking yields the best overall performance (Figs.~\ref{fig:heatmap} and~\ref{fig:swarm}). RF stacking provides high accuracy on most networks for distinguishing missing links from non-edges (AUC) and moderate accuracy on maximizing accurate predictions on a budget (Top-k), and is very fast to train using only low-cost (CPU) hardware. That is, RF stacking is a scalable, low-cost, high-accuracy solution to the LP problem, particularly if the goal is distinguishing missing links from non-edges.
If hardware costs are not a constraint, GraphSAGE performs well (Figs.~\ref{fig:heatmap} and~\ref{fig:swarm}), achieving moderate accuracy on most networks under AUC and high accuracy on Top-k, needing moderate training time on high-cost (GPU) hardware. That is, GraphSAGE is a less-scalable, higher-cost, but high-accuracy solution to the LP problem, particularly if the goal is accurate predictions on a budget. In general, graph embedding models like GCN and SAGE perform better on Top-k predictions compared to stacking algorithms, but not always.
Table~\ref{tab:qualitative_comp} summarizes our assessments of all algorithms across four dimensions:\ training speed, hardware cost, AUC accuracy, and Top-k accuracy.

%%%%% TABLE 6 : GENERAL ASSESSMENT OF ALGORITHMS
\begin{table}[t!]
\setlength{\tabcolsep}{3pt} % Adjust column spacing
\renewcommand{\arraystretch}{1.3} % Adjust row spacing
\centering
%\begin{adjustbox}% Fit table to one column
{\small % Larger font size
\begin{tabular}{|l|c|c|c|c|}
\hline
\vspace{-0.5em} \textbf{Algorithm}          & \textbf{Training} & \textbf{Hardware} & 
\textbf{AUC} & \textbf{Top-k}  \\ 
                         & \textbf{Speed}   & \textbf{Cost}              & 
\textbf{Accuracy}     & \textbf{Accuracy} \\ \hline

RF          & \textcolor[rgb]{0.0,0.5,0.0}{Fast} & \textcolor[rgb]{0.0,0.5,0.0}{Low} & \textcolor[rgb]{0.0,0.5,0.0}{High} & \textcolor{blue}{Moderate} \\

\hline
XGB                 & \textcolor[rgb]{0.0,0.5,0.0}{Fast} & \textcolor[rgb]{0.0,0.5,0.0}{Low} & \textcolor{blue}{Moderate}    & \textcolor{blue}{Moderate}       \\ \hline

LR                 & \textcolor[rgb]{0.0,0.5,0.0}{Fast} & \textcolor[rgb]{0.0,0.5,0.0}{Low} & \textcolor{red}{Low}    & \textcolor{blue}{Moderate}       \\

\hline
SVM                     & \textcolor{red}{Slow}   & \textcolor[rgb]{0.0,0.5,0.0}{Low} & \textcolor{blue}{Moderate}    & \textcolor{red}{Low} \\ 

\hline
GCN             & \textcolor[rgb]{0.0,0.5,0.0}{Fast} & \textcolor{red}{High}  & \textcolor{blue}{Moderate}    & \textcolor{blue}{Moderate}                                \\ 

\hline
SAGE               & \textcolor{blue}{Moderate} & \textcolor{red}{High} & \textcolor{blue}{Moderate} & \textcolor[rgb]{0.0,0.5,0.}{High}    \\ \hline
\hline
\hline
meta-learn (restr.)               & \textcolor{blue}{Moderate} & \textcolor{red}{High} & \textcolor[rgb]{0.0,0.5,0.0}{High} & \textcolor[rgb]{0.0,0.5,0.}{High}    \\ 
\hline
meta-learn (full)               & \textcolor{red}{Slow*} & \textcolor{red}{High} & \textcolor[rgb]{0.0,0.5,0.0}{High} & \textcolor[rgb]{0.0,0.5,0.}{High}    \\ \hline
\end{tabular}
} % End larger font
\vspace{2mm}\caption{General assessment of six state-of-the-art algorithms and two meta-learning algorithms (restricted chooses best algorithm from RF stacking or SAGE; full chooses best algorithm from all six), along four dimensions:\ training speed, hardware cost, accuracy at distinguishing missing links from non-edges (AUC), and accuracy for predictions on a budget (Top-k). If hardware cost is a constraint, then RF stacking is the best performing model, and if hardware casts are not limiting, then the restricted meta-learner performs better.}
%\end{adjustbox}
\label{tab:qualitative_comp}
\end{table}
%%%%%%%%%%%%%%%%%%%%%%%%%%%%%%%%%%%

% mean time per network
% Model                            Time 
% XGB                              0.434969
% LR                               0.869387
% SAGE                             2.233884
% GNN                              0.729680
% RF                               1.124787
% SVM                            140.905318
% model1_Top_pred_time            24.467500
% model2_Top_pred_time             2.119971
% model_1_best_model_AUC_time     15.037767
% model_2_best_model_AUC_time      1.583397

Exploiting the dependence of algorithm performance on network structure, the meta-learner algorithm we introduce can predict which particular state-of-the-art algorithm will yield the most accurate predictions for an input network, given easily calculated characteristics of it. The full meta-learner (select from all six algorithms) achieves a substantial improvement over the RF stacking and SAGE baselines in both AUC and Top-k, indicating that optimal link prediction is achieved by learning to use the best algorithm for a given network. A restricted meta-learner (select from RF stacking or SAGE only) is slightly less accurate than the full meta-learner. Both meta-learners suffer a slight penalty to their accuracy by having to learn from data, and our results suggest that better network features could potentially further improve these meta-learners' accuracies for the LP problem. We include in Table~\ref{tab:qualitative_comp} our assessment of these meta-learner algorithms across the four dimensions of training speed, hardware cost, AUC accuracy, and Top-k accuracy. One key question for future work is how well these supervised techniques perform under alternative choices of the missingness function~\cite{aiyappa2024degreebias,he2024sampling} that defines which edges are observed and which are missing.

The high accuracy across algorithms on social networks (about 90\% of the time, an algorithm's accuracy will be within 3\% of the best algorithm for that network) suggests that prior work on LP algorithms may be implicitly over-fitted toward the structure of these networks, and the substantially lower accuracies on non-social networks may indicate that more work is needed to improve predictions in these harder settings. At the same time, however, we discover there exists some social networks for which all LP algorithms still perform poorly. Future work should investigate what makes these social networks harder than most other social networks.

These findings highlight the importance of choosing and applying the right algorithm for each particular network to predict its missing links, rather than assuming that one class of LP algorithms are always best, e.g., graph neural networks. These findings also highlight the flexibility, scalability, and high performance of meta-learning algorithms for the LP problem, which efficiently help solve the problem of choosing which algorithm to apply in practice to any particular network.

\bigskip
\noindent \textbf{Code and Data Availability:} A Python implementation of the meta-learner algorithm and all other LP algorithms used in this study, can be found at  \url{https://github.com/bis1999/edgepredict}.
The \textit{OptimalLinkPrediction} benchmark data set of networks was obtained from \url{https://github.com/Aghasemian/OptimalLinkPrediction}.

\bigskip
\noindent \textbf{Acknowledgements:} This work was supported in part by National Science Foundation Award IIS 1956183 (BS, AC). The authors acknowledge the BioFrontiers Computing Core at the University of Colorado Boulder for providing High Performance Computing resources supported by BioFrontiers IT.


% macro to make Figures numbered as S#
\renewcommand{\thefigure}{S\arabic{figure}}
\setcounter{figure}{0}
% macro to make Tables numbered as S#
\renewcommand{\thetable}{S\arabic{table}}
\setcounter{table}{0}

\appendix

\section*{Appendix}

%%% APPENDIX A
\section{Link Prediction Benchmark Corpus}\label{appendix:corpus}

We use the previously published \textit{OptimalLinkPrediction} benchmark: \url{ https://github.com/Aghasemian/OptimalLinkPrediction}. This benchmark contains 550 structurally diverse networks that are drawn from six scientific domains~\cite{ghasemian2020stacking}:\ biological (179), social (124), transportation (35), technological (70), economic (124), and informational (18). This benchmark has been used in multiple past studies of LP algorithms and is available publicly. All networks in this benchmark are simple graphs (edges are undirected, unweighted, and there are no self-loops), none are bipartite, and none have node or edge attributes.

Table~\ref{tab:network_stats} provides an overview of the contents of the benchmark, and Figure~\ref{fig:corpus} shows the diverse range of mean degrees and sizes across scientific domains.

%%%%%%%%%%% TABLE 4: summary of LP benchmark data set
\begin{table}[h!]
\centering
\resizebox{\columnwidth}{!}{%
\begin{tabular}{l|c|c|c|c}
\textbf{\textcolor{black}{Network Domain}} & \textbf{\textcolor{black}{$\langle n \rangle$ }} & \textbf{\textcolor{black}{$\langle m \rangle$ }} & \textbf{\textcolor{black}{$\langle k \rangle$}} & \textbf{\textcolor{black}{Net. Count}} \\ \hline
\textbf{Full Corpus}                       & 509.75                                                       & 1154.92                                                     & 5.42                                      & 550                                     \\ \hline \hline
Biological                                 & 294.23                                                       & 780.15                                                      & 6.30                                      & 179                                     \\ \hline
Economic                                   & 701.68                                                       & 865.69                                                      & 3.34                                      & 124                                     \\ \hline
Informational                              & 494.17                                                       & 1266.22                                                     & 5.32                                      & 18                                      \\ \hline
Social                                     & 558.58                                                       & 1988.33                                                     & 7.59                                      & 124                                     \\ \hline
Technological                              & 532.54                                                       & 1061.03                                                     & 4.03                                      & 70                                      \\ \hline
Transportation                             & 721.34                                                       & 1274.14                                                     & 3.49                                      & 35                                      \\ \hline
\end{tabular}%
}
\caption{Summary of the contents of the LP benchmark used in this study, showing broad representation across six scientific domains.}
\label{tab:network_stats}
\end{table}
%%%%%%%%%%%%%%%%%%%%%%%%%%

%%% APPENDIX B
\section{Choices for Stacking Algorithm} \label{appendix:level1}
We consider four different choices for the level-1 meta-learner algorithm in the stacked models we evaluate. Because these are standard supervised learning algorithms, we provide only a very brief overview of them here.

\textit{Random Forests} (RF) are an ensemble method that constructs multiple decision trees to improve prediction accuracy and avoid overfitting. It achieves diversity by bootstrapping the data for each tree and selecting a random subset of features for splits. Key hyperparameters in random forests include the number of trees, their depth, and the number of features considered for each split. Tuning these parameters can significantly influence the model's performance and its ability to generalize to new data~\cite{Breiman2001}.

\textit{XGBoost} (XGB) sequentially builds decision trees, each correcting errors made by the previous ones through gradient boosting. XGBoost optimizes a differentiable loss function with core hyperparameters, including the learning rate, the maximum depth of the trees, and the number of trees. Adjusting these parameters helps control the model's complexity and speed of convergence~\cite{Chen2016}.

\textit{Logistic Regression} (LR), in contrast, is a probabilistic algorithm for learning a linear combination of features to solve binary classification problems, e.g., whether a link exists or not, and is widely used in science and engineering applications. The main hyperparameters include the regularization strength and the type of penalty (e.g., L1 or L2 regularization), which help prevent overfitting by penalizing large coefficients~\cite{Hosmer2013}.

\textit{Support Vector Machines} (SVM) are a commonly used algorithm for classification and regression problems. SVMs aim to find an optimal separating hyperplane between classes by maximizing the margin, the distance between the closest points of the classes (support vectors). Key hyperparameters include the penalty parameter~$C$, which controls the trade-off between achieving a low error on the training data and minimizing the norm of the coefficients, and the kernel type, which determines the transformation of the input data space~\cite{Cortes1995}.

%%% APPENDIX C
\section{List of Topological Predictors for Stacking}\label{appendix:level0}

\begin{table*}[t!]\addtolength{\tabcolsep}{-5pt}
\caption{Abbreviations and descriptions of 42 topological predictors, across three types:\ \textit{global} predictors (7), which are functions of the entire network and whose utility is in providing context to other predictors; \textit{pair-level} predictors (15), which are functions of the joint topological properties of the pair $i,j$; and \text{node-level} predictors (20), which are functions of the independent topological properties of the nodes $i$ and $j$, producing one value for each node in the pair $i,j$.}
\vspace*{-3mm}
\centering
 \begin{tabular}{|p{2.42cm} | p{10.6cm} |c|c|c|c|} 
 
 \hline
Abbreviation &  Description & \hspace{1mm}Global\hspace{1mm} & \hspace{1mm}Pair\hspace{1mm} & \hspace{1mm}Node\hspace{1mm} & ~~Ref.~~  \\ [0.5ex] 
 \hline\hline
 N & number of nodes & $\bullet$ & & & \cite{newman:networks:2018} \\ \hline
 OE & number of observed edges & $\bullet$ & & & \cite{newman:networks:2018} \\ \hline
 AD & average degree & $\bullet$ & & & \cite{newman:networks:2018} \\ \hline
 VD & variance of degree distribution & $\bullet$ & & & \cite{newman:networks:2018} \\ \hline
 ND & network diameter & $\bullet$ & & & \cite{newman:networks:2018} \\ \hline
 DA & degree assortativity of graph & $\bullet$ & & & \cite{networkx} \\ \hline
 NT & network transitivity (clustering coefficient) & $\bullet$ & & & \cite{newman:networks:2018} \\ \hline
 ACC & average (local) clustering coefficient & $\bullet$ & & & \cite{newman:networks:2018} \\ [0.5ex]  \hline\hline
CN & common neighbors of $i,j$ & & $\bullet$ & & \cite{liben2007link} \\ \hline
SP & shortest path between $i,j$ & & $\bullet$ & & \cite{liben2007link} \\ \hline
LHN & Leicht-Holme-Newman index of neighbor sets of $i,j$& & $\bullet$ & & \cite{leicht2006vertex} \\ \hline
PPR & $j$-th entry of the personalized page rank of node $i$ & & $\bullet$ & & \cite{networkx} \\ \hline
PA & preferential attachment (degree product) of $i,j$ & & $\bullet$ & & \cite{liben2007link} \\ \hline
JC & Jaccard's coefficient of neighbor sets of $i,j$ & & $\bullet$ & & \cite{liben2007link} \\ \hline
AA & Adamic/Adar index of $i,j$ & & $\bullet$ & & \cite{liben2007link} \\ \hline
RA & resource allocation index of $i,j$ & & $\bullet$ & & \cite{networkx} \\ \hline
LRA & entry $i,j$ in low rank approximation (LRA) via singular value decomposition (SVD) & & $\bullet$ & & \cite{cukierski2011graph} \\ \hline
dLRA & dot product of columns $i$ and $j$ in LRA via SVD for each pair of nodes $i,j$ & & $\bullet$ & & \cite{cukierski2011graph} \\ \hline
mLRA & average of entries $i$ and $j$'s neighbors in low rank approximation & & $\bullet$ & & \cite{cukierski2011graph} \\ \hline
LRA-approx & an approximation of LRA & & $\bullet$ & & \cite{cukierski2011graph} \\ \hline
dLRA-approx & an approximation of dLRA & & $\bullet$ & & \cite{cukierski2011graph} \\ \hline
mLRA-approx & an approximation of mLRA & & $\bullet$ & & \cite{cukierski2011graph} \\ [0.5ex]  \hline\hline
LCC$_i$, LCC$_j$ & local clustering coefficients for $i$ and $j$ & &  & $\bullet$ & \cite{networkx} \\ \hline
AND$_i$, AND$_j$ & average neighbor degrees for $i$ and $j$ & &  & $\bullet$ & \cite{networkx} \\ \hline
SPBC$_i$, SPBC$_j$ & shortest-path betweenness centralities for $i$ and $j$ & &  & $\bullet$ & \cite{networkx} \\ \hline
CC$_i$, CC$_j$ & closeness centralities for $i$ and $j$ & &  & $\bullet$ & \cite{networkx}  \\ \hline
DC$_i$, DC$_j$ & degree centralities for $i$ and $j$ & &  & $\bullet$ & \cite{networkx} \\ \hline
EC$_i$, EC$_j$ & eigenvector centralities for $i$ and $j$ & &  & $\bullet$ & \cite{networkx} \\ \hline
KC$_i$, KC$_j$ & Katz centralities for $i$ and $j$ & &  & $\bullet$ & \cite{networkx} \\ \hline
LNT$_i$, LNT$_j$ & local number of triangles for $i$ and $j$ & &  & $\bullet$ & \cite{networkx} \\ \hline
PR$_i$, PR$_j$ & Page rank values for $i$ and $j$ & &  & $\bullet$ & \cite{networkx} \\ \hline
LC$_i$, LC$_j$ & load centralities for $i$ and $j$ & &  & $\bullet$ & \cite{networkx} \\ \hline
 \end{tabular}
 \label{table:top_feat}
\end{table*}

The level-0 predictors we use are functions of the topological properties of the target pair of nodes~$i,j$ being evaluated. Each of these topological predictors has been used in prior work on the LP problem, and this particular set replicates the feature set used in Ref.~\cite{ghasemian2020stacking}. Topological predictors are divided into three classes:\ global, pairlevel predictors, and node-level predictors. Table~\ref{table:top_feat} provides a complete list of these predictors.
\\

\noindent \textbf{Global predictors.} 
These 8 predictors are network-level features that provide context to other predictors. They include the number of nodes (N); the number of observed edges (OE); the average degree (AD); the variance of the degree distribution (VD); the network diameter (ND); the degree assortativity of the network (DA); the network transitivity (clustering coefficient) (NT); and the average local clustering coefficient (ACC).
\\

\noindent\textbf{Pair-level predictors.}~~
Pair-level or pairwise predictors encompass 14 features:\ the number of common neighbors of $i,j$ (CN); the length of the shortest path between $i,j$ (SP); the Leicht-Holme-Newman index of neighbor sets of $i,j$ (LHN); the personalized page rank (PPR) of~$j$ from~$i$; the degree product or `preferential attachment' of $i,j$ (PA); the Jaccard coefficient of the neighbor sets of $i,j$ (JC); Adamic-Adar index of $i,j$ (AA), the resource allocation index of $i,j$ (RA); the entry $i,j$ of a low rank approximation (LRA) via a singular value decomposition; the dot product of the $i,j$ columns in the LRA via singular value decomposition (dLRA); the mean of entries~$i$ and~$j$'s neighbors in the LRA (mLRA); and simple approximations of the latter three predictors (LRA-approx, dLRA-approx, mLRA-approx)~\cite{newman:networks:2018,liben2007link, al2011link, cukierski2011graph} .
\\

\noindent \textbf{Node-level predictors.} 
Node-level predictors encompass 10 functions of the independent topological properties of the individual nodes $i$ and $j$, producing a pair of predictor values. Unlike the pairwise predictors, which are themselves unsupervised algorithms to predict missing links, these node-level predictors do not directly score the likelihood that $i,j$ is a missing link. Rather, these features require a supervised framework to learn how to convert the particular pair of values into a score.

The 20 node-based predictors are two instance each of:\ the local clustering coefficient (LCC); the average neighbor degree (AND); the shortest-path betweenness centrality (SPBC); the closeness centrality (CC); the degree centrality (DC); the eigenvector centrality (EC); the Katz centrality (KC); the local number of triangles (LNT); the Page rank (PR); and the load centrality (LC)~\cite{newman:networks:2018,liben2007link, al2011link, cukierski2011graph}.

%%% APPENDIX E
\subsection{GCN and GraphSAGE Link Predictions}
This section provides a mathematical overview of our link-prediction framework using Graph Convolutional Networks (GCNs)~\cite{kipf2017semisupervised} and GraphSAGE~\cite{hamilton2018inductive}, focusing on how node embeddings are computed and used to predict edges. Note that we do not use any \emph{external} feature matrix for nodes; instead, each node is initialized with a learnable embedding.
\\

\noindent\textbf{Feature Matrix and Node Embeddings}
\label{sec:features}

Let the graph be $G=(V, E)$ with $|V|=n$ nodes and $|E|$ edges, where each edge $(i,j)\in E$ indicates a positive (actual) link in the graph. In many graph neural network (GNN) settings, each node $v\in V$ has a feature vector $x_v \in \mathbb{R}^d$, forming a feature matrix $\mathbf{X}\in \mathbb{R}^{n\times d}$. However, in our experiments, we do \emph{not} use external node features. Instead, we initialize a node embedding matrix 
\begin{equation}
H^{(0)} \sim \mathcal{U}\Bigl(-\tfrac{1}{\sqrt{k}},\,\tfrac{1}{\sqrt{k}}\Bigr),
\label{eq:xavier}
\end{equation}
where $H^{(0)}\in\mathbb{R}^{n\times k}$ is learned via Xavier initialization, and $k$ is the embedding dimension. These embeddings are \emph{optimized} during training to improve link prediction accuracy.
\\

\noindent\textbf{Graph Convolutional Network (GCN)}
\label{sec:gcn}

A GCN~\cite{kipf2017semisupervised} computes new embeddings by normalizing and aggregating each node's neighbors. Let $h_v^{(\ell)} \in \mathbb{R}^{d_{\ell}}$ denote the embedding of node $v$ at layer $\ell$, and let $\mathcal{N}(v)$ be the set of $v$'s neighbors. The GCN update from layer $\ell$ to $\ell+1$ is:
\begin{equation}
h_v^{(\ell+1)} 
~=~ 
\sigma\Bigl(\!
  \sum_{u \in \mathcal{N}(v)\cup\{v\}} 
    \tfrac{1}{\sqrt{d_u\,d_v}}\,W^{(\ell)}\,h_u^{(\ell)}
\Bigr),
\label{eq:gcn_update}
\end{equation}
where $W^{(\ell)}$ is a trainable weight matrix, $\sigma(\cdot)$ is a non-linear activation (e.g., ReLU), and $d_u$ (resp.\ $d_v$) denotes the degree of node $u$ (resp.\ $v$). After each layer, we apply dropout for regularization.

In matrix form, \eqref{eq:gcn_update} can be written as:
\begin{equation}
\mathbf{H}^{(\ell+1)} 
~=~
\widetilde{\mathbf{D}}^{-\tfrac12}
\,\widetilde{\mathbf{A}}
\,\widetilde{\mathbf{D}}^{-\tfrac12}
\;\mathbf{H}^{(\ell)}\,W^{(\ell)},
\label{eq:gcn_matrix}
\end{equation}
where $\widetilde{\mathbf{A}} = \mathbf{A} + \mathbf{I}$ includes self-loops, and $\widetilde{\mathbf{D}}$ is the diagonal degree matrix of $\widetilde{\mathbf{A}}$.
\\

\vspace{8mm}\noindent\textbf{GraphSAGE (SAGE)}
\label{sec:sage}

GraphSAGE~\cite{hamilton2018inductive} was designed for large-scale graphs by \emph{sampling} a fixed number of neighbors per node at each layer. Formally, a random subset $\widetilde{\mathcal{N}}(v) \subseteq \mathcal{N}(v)$ is often used to reduce computation:
\begin{equation}
\widetilde{\mathcal{N}}(v) 
~=~
\text{RandomSample}\bigl(\mathcal{N}(v), k\bigr),
\end{equation}
where $k$ is a chosen sample size. In our setting, we \emph{do not} subsample; instead, we use the entire neighborhood. Hence, letting $h_v^{(\ell)}$ be the embedding of node $v$ at layer $\ell$, the \emph{mean} aggregation step is:
\begin{equation}
h_v^{(\ell+1)} 
~=~
\sigma\Bigl(
  W^{(\ell)}
  \cdot
  \mathrm{MEAN}\!\bigl(\{\,h_v^{(\ell)}\} \cup \{\,h_u^{(\ell)}: u \in \mathcal{N}(v)\}\bigr)
\Bigr),
\label{eq:sage_mean}
\end{equation}
where
\[
\mathrm{MEAN}\bigl(\{\,h_v^{(\ell)}\}\cup \{\,h_u^{(\ell)}: u \in \mathcal{N}(v)\}\bigr)
~=~
\tfrac{1}{|\mathcal{N}(v)| + 1}
\sum_{u\in \mathcal{N}(v)\cup \{v\}}
  h_u^{(\ell)}.
\]
We thus capture \emph{all} neighbors at every step. Although this can be more expensive for high-degree nodes, it yields potentially higher accuracy.
\\

\noindent\textbf{Link Prediction Model}
\label{sec:linkpred}

After $L$ layers of either GCN or GraphSAGE, we obtain final node embeddings $h_i^{(L)}, h_j^{(L)}$ for each node pair $(i,j)$. To predict whether $(i,j)$ is an edge, we form an element-wise product:
\begin{equation}
z_{ij} 
~=~
h_i^{(L)} \,\odot\, h_j^{(L)} \enspace ,
\label{eq:elemwise_mult}
\end{equation}
and pass it through fully-connected layers (an MLP). Concretely, let $z_{ij}^{(0)} = z_{ij}$ and define:
\begin{equation}
z_{ij}^{(\ell+1)} 
~=~
\sigma\bigl(
  W_{\mathrm{FC}}^{(\ell)}\,z_{ij}^{(\ell)} 
  \;+\;
  b^{(\ell)}
\bigr) \enspace .
\end{equation}
The final layer applies a sigmoid:
\begin{equation}
\hat{y}_{ij}
~=~
\text{sigmoid}\bigl(z_{ij}^{(L')}\bigr) \enspace ,
\label{eq:lp_prediction}
\end{equation}
yielding the predicted probability that $(i,j)$ is an edge.
\\

\noindent\textbf{Training and Optimization}
\label{sec:training}

We train by sampling:
\begin{itemize}
    \item \emph{Positive edges:} $(i,j)\in E$.
    \item \emph{Negative edges:} $(i,j)\in E_{\mathrm{neg}}$, where $E_{\mathrm{neg}}$ is a sampled subset of node pairs not in $E$ [i.e., $(i,j)\notin E$].
\end{itemize}
The binary cross-entropy (BCE) loss consists of two terms, one for positive edges and one for negative edges:
\begin{equation}
\label{eq:bce_loss}
\mathcal{L} 
~=~
-\frac{1}{|E|}
  \sum_{(i,j)\in E}
    \log\bigl(\hat{y}_{ij} + \varepsilon\bigr)
~-\frac{1}{|E_{\mathrm{neg}}|}
  \sum_{(i,j)\in E_{\mathrm{neg}}}
    \log\bigl(1 - \hat{y}_{ij} + \varepsilon\bigr),
\end{equation}
where $\varepsilon$ is a small constant (e.g., $10^{-15}$) preventing numerical issues. We update all parameters
\[
\theta ~\in~ \bigl\{\,W^{(\ell)},\,W_{\mathrm{FC}}^{(\ell)},\,H^{(0)}\bigr\}
\]
using Adam:
\begin{equation}
\theta 
\;\leftarrow\;
\theta 
~-~
\eta
\,\nabla_{\theta}
\,\mathcal{L} \enspace ,
\end{equation}
with gradient clipping ($\|\nabla\|\le \tau$) to maintain stability.
\\

\noindent\textbf{Validation and Prediction}
\label{sec:validation}

After training, we produce the final node embeddings $H^{(L)}=\text{Model}(H^{(0)},A)$ and compute edge scores via \eqref{eq:elemwise_mult}--\eqref{eq:lp_prediction}. We measure performance using the standard AUC metric:
\begin{equation}
\mathrm{AUC} 
~=~
\frac{1}{|P|\cdot |N|}
\sum_{(i,j)\in P}\sum_{(k,\ell)\in N}
 \mathbf{1}\bigl(\hat{y}_{ij} \;>\;\hat{y}_{k\ell}\bigr) \enspace ,
\label{eq:auc}
\end{equation}
where $P$ and $N$ are sets of positive and negative edges in the validation (or test) split.

%%% APPENDIX D
\section{Hyper Parameter Tuning}\label{appendix:hyper}
\subsection{Stacking Algorithms}

As stated in our experimental design, we used the network $G''$ to calculate the topological features for pairs of true edges and missing edges. Once we had created this dataset, we used SMOTE~\cite{chawla2002smote} (Synthetic Minority Oversampling Technique) to address class imbalances in the dataset. The upsampling is an important part of training as we know we have a small number of true edges compared to missing edges. The selection of hyperparameters was conducted through a 5-fold grid-search. Basically, we use a dataset created after upsampling, and use 20\% of pairs as testing pairs for each folder in our cross-validation. The primary evaluation metric for hyperparameter selection was AUC.

\subsection{Graph Neural Network Algorithms}
Hyperparameter tuning for the graph neural network algorithms (GCN, SAGE) optimizes the number of layers and the dimensionality of the embeddings. 

The initial step involves constructing the adjacency matrix based on the observed network and initializing feature vectors with Gaussian weights, as we do not leverage any node attributes~\cite{kipf2017semisupervised}. Training these models entails providing them with positive and negative examples and utilizing a log loss function~\cite{kipf2017semisupervised}. The training process for these models is then divided into two stages. First, we optimize parameters like the number of dimensions and hidden layers. This is done by creating the adjacency matrix based on $G''$ and using $E''$ edges as positive samples, while the remaining edges serve as negatives. This occurs within a cross-validation framework. Subsequently, we evaluate the model's performance on the unseen edges $|E' - E''|$ and negative samples. Once optimal parameters are determined, we construct the adjacency matrix on the network $G'$ and train the models using the positive edges $|E'|$. We then assess the model's ability to differentiate between the holdout set and the remaining negative samples. Potential overfitting during model evaluation, particularly due to the inclusion of negative samples during training, may slightly bias performance evaluation.

\begin{table*}[t] % Force top placement
    \addtolength{\tabcolsep}{-5pt}
    \caption{Hyperparameter selection details for machine learning models. The hyperparameter tuning was performed using \texttt{scikit-learn}~\cite{scikit-learn} for traditional models and \texttt{PyTorch Geometric}~\cite{Fey/Lenssen/2019} for GCN and SAGE.}
    \vspace*{-3mm}
    \centering
    \begin{tabular}{|p{3cm}|p{12cm}|p{2.5cm}|} 
        \hline
        \textbf{Model} & \textbf{Hyperparameters} & \textbf{Ref} \\
        \hline\hline
        RF & max\_depth: \{3, 6, 9\}, n\_estimators: \{25, 50, 100, 125\}, criterion: gini & \cite{scikit-learn} \\
        \hline
        LR & C: \{0.1, 1, 10, 100\}, penalty: \{L1, L2\}, solver: \{liblinear, saga\}, max\_iter: 1000 & \cite{scikit-learn} \\
        \hline
        XGB & n\_estimators: \{50, 100, 200\}, max\_depth: \{5, 10, 25, 50\}, learning\_rate: 0.1, subsample: 1.0, colsample\_bytree: 0.7, gamma: 0.5 & \cite{scikit-learn} \\
        \hline
        SVM & C: \{0.1, 1, 10, 100\}, kernel: rbf, gamma: scale & \cite{scikit-learn} \\
        \hline
        GCN & num\_layers: \{2, 3, 4\}, hidden\_channels: \{32, 64, 128\}, dropout: 0.5, optimizer: Adam, learning\_rate: 0.005, activation: ReLU & \cite{Fey/Lenssen/2019} \\
        \hline
        SAGE & num\_layers: \{2, 3, 4\}, hidden\_channels: \{32, 64, 128\}, dropout: 0.5, optimizer: Adam, learning\_rate: 0.005, activation: ReLU, aggregation: mean & \cite{Fey/Lenssen/2019} \\
        \hline
    \end{tabular}
    \label{tab:hyperparams}
\end{table*}

%%% APPENDIX F
\section{LP algorithm performance correlates with network structure}\label{appendix:Trend}

To characterize the dependence of LP algorithm performance on variation in network structure, we measure how AUC and Top-k accuracy changes for all six state-of-the-art algorithms as a function of four network measures:\
\begin{enumerate}
    \itemsep-0.1pt
    \item the mean degree $\langle k \rangle$;
    \item degree assortativity $r$, measuring the tendency for nodes with similar degrees to be connected;
    \item mean local clustering $\langle C_i \rangle$, measuring the density of a node' immediate neighborhood; and 
    \item the mean geodesic (shortest) path length $\langle \ell \rangle$ among node pairs.
\end{enumerate}

\begin{figure*}[t!]
\centering
\begin{tabular}{cc}
\includegraphics[width=1\textwidth]{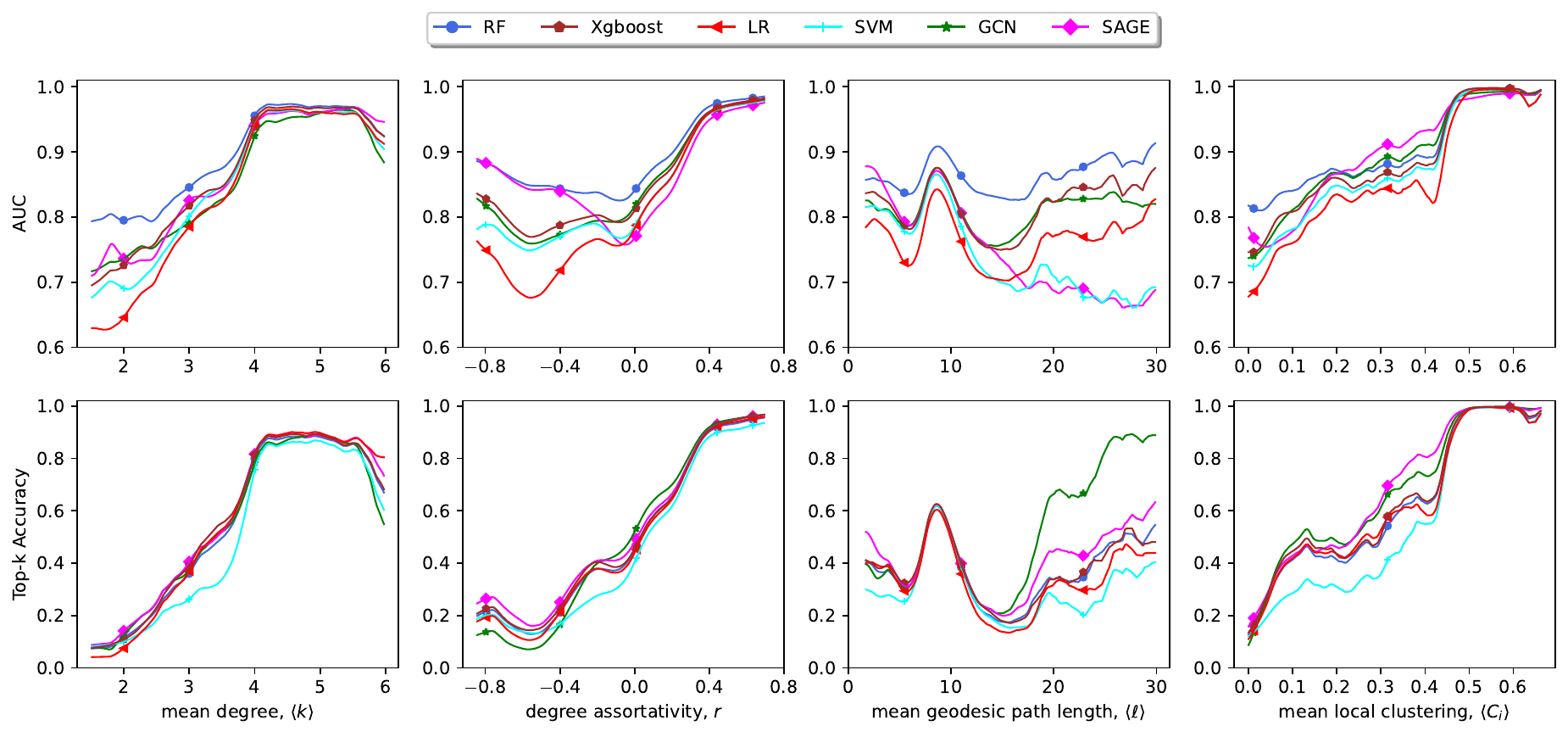}
\end{tabular}
\vspace{-4mm}
\caption{Link prediction accuracy under two measures of accuracy (top row:\ AUC; bottom row:\ Top-k) for six state-of-the-art algorithms (RF, XGB, LR, and SVM stacking, and GCN and SAGE graph neural networks) as a function of four measures of network structure:\ mean degree $\langle k\rangle$, degree assortativity $r$, mean geodesic path length $\langle \ell \rangle$, and mean local clustering coefficient $\langle C_i\rangle$. The patterns indicate that accuracy generally improves with larger mean degree, most positive degree assortativity, and more local clustering (see Appendix~\ref{appendix:Trend}.}
\label{fig:Trend}
\end{figure*}

Figure~\ref{fig:Trend} shows the results. To reveal the underlying trends, to each data series we apply a weak exponential smoother. All algorithms generally exhibit improved accuracy with increasing mean degree $\langle k \rangle$, degree assortativity $r$, mean local clustering $\langle C_i \rangle$, and mean geodesic path length $\langle \ell \rangle$. More specifically, more edges (higher $\langle k \rangle$) and denser neighborhoots (higher $\langle C_i \rangle$) strongly correlate with improved accuracy for both distinguishing missing links from non-edges (AUC) and accuracy predictions on a budget (\mbox{Top-k}). Greater degree assortativity $r$ improves Top-k accuracy for all algorithms, but only when $r>0$ for AUC. In contrast, when $r<0$, we see algorithm-specific performance differences, with SAGE and RF stacking performing best with degree disassortative structure ($r<0$). Mean geodesic path length $\langle \ell \rangle$ presents a more complicated pattern:\ for very compact networks ($\langle \ell \rangle<15$), algorithms perform similarly under Top-k, while for less compact networks ($\langle \ell \rangle>15$), we see algorithm-specific performance differences, with GCN performing best, followed by SAGE, and then RF stacking and XGB stacking. Under AUC, we find less dependence overall with mean path length $\langle \ell \rangle$, but with RF stacking performing best at all values, followed by XGB stacking and GCN.

Overall, these results indicate that the accuracy of RF stacking is relatively high regardless of variations in network topology, particularly under AUC. And, SAGE exhibits similar robustness under Top-k. Notably, in very sparse networks, with $\langle k \rangle < 4$, RF stacking outperforms all other models in AUC.

% -------- FIGURE 5 -----------
% -------------------------

\section{Regression Model Training and Feature Importance Analysis}\label{appendix:regmod}
We train two regression models to predict, from easily calculated measures of network structure, the best link prediction performance (across all algorithms) in terms of either AUC and Top-k accuracy. 
The feature set is composed of six topological features:\  the mean local clustering coefficient $\langle C_i\rangle$, mean geodesic path length $\langle \ell \rangle$, degree assortativity $r$, number of nodes $n$, mean degree $\langle k\rangle$, and variance of degrees $\sigma^2_{k}$. The dataset is split into training and test sets based on unique network indices, ensuring that different topological variants of the same network remain in the same split to prevent data leakage. In each iteration, $80\%$ of the networks are used for training, and $20\%$ are held out for testing.

A Random Forest Regressor is used to model the relationship between network features and link prediction performance, with separate models for predicting AUC and Top-k accuracy. This entire process is repeated 100 times using different random splits to ensure robustness and measure variation in estimated feature importance. Model performance is evaluated using the $r^2$ score (coefficient of determination), which measures how well the model explains the variance in the target variable. The average $r^2$ across all runs provides an estimate of the model's predictive power.

To understand how each features contributes to the model, we compute the Gini importance (mean decrease in impurity) for each feature, after training. We average the estimated importance values across all 100 runs to produce stable feature rankings. This allows us to identify the most critical topological metrics driving link prediction performance, offering interpretable insights into how network structure influences AUC and Top-k accuracy.

%%% APPENDIX G
\section{Meta-learning Model Training}\label{appendix:meta}
The classifier is trained to predict the best-performing link prediction model based on network topology features using an XGBoost classifier.

The dataset comprises the same topological features as used to predict the best model performance (Appendix~\ref{appendix:regmod}):\ the mean local clustering coefficient $\langle C_i\rangle$, mean geodesic path length $\langle \ell \rangle$, degree assortativity $r$, number of nodes $n$, mean degree $\langle k\rangle$, and variance of degrees $\sigma^2_{k}$, extracted from network $G'$, along with performance metrics (AUC and Top-k accuracy) computed for multiple link prediction models. To define classification labels, we determine the best-performing model for each sample by selecting the model with the highest AUC or Top-k score. The classification tasks are structured as follows:
\begin{itemize}
    \item \textbf{Model 1} (multi-class classification):\\ Predict the best-performing model among six candidates:\ \textit{RF}, \textit{XGB}, \textit{LR}, \textit{SVM}, \textit{GCN}, and \textit{SAGE}.
    \item \textbf{Model 2} (binary classification):\\ Predict the best-performing model between \textit{RF} and \textit{SAGE}.
\end{itemize}

Each classifier is trained using an 80-20 train-test split, ensuring that all topological variants of a given network remain within the same split to prevent data leakage. Hyperparameter tuning is performed via grid-search over parameters such as maximum depth, number of estimators, subsampling rate, and column sampling rate. The best model is selected based on accuracy scores using three-fold cross-validation. 

Predictions are evaluated on the test set by comparing the classifier's predicted best model against the actual best-performing model based on AUC or Top-k accuracy. To ensure robustness, this process is repeated 100 times, storing both predicted and actual performances for further analysis. This iterative approach enables a thorough assessment of the classifier's generalization capability across different network structures and provides insights into the influence of topological features on link prediction performance.

\end{document}